\documentclass[
    aps,
    prr,
    twocolumn,
    superscriptaddress,
    amsmath,amssymb,
]{revtex4-2}

\usepackage{graphicx}% Include figure files
\usepackage{bm}% bold math
\usepackage{xr} % cross-referencing
\usepackage[breaklinks=true]{hyperref}

\hypersetup{%
  colorlinks=true,%
  linkcolor=blue,%
  citecolor=blue,%
  urlcolor=black,%
}

\newtheorem{thm}{Theorem}
\newtheorem{cor}[thm]{Corollary}

\begin{document}

%\preprint{}

\title{Thermodynamic Limit in Learning Period Three}

\author{Yuichiro Terasaki}
\email[]{terasaki@isi.imi.i.u-tokyo.ac.jp}
%\thanks{}
% \altaffiliation{}
\affiliation{Graduate School of Information Science and Technology, The University of Tokyo, Tokyo 113-8656, Japan}

\author{Kohei Nakajima}
\email[]{k-nakajima@isi.imi.i.u-tokyo.ac.jp}
%\thanks{}
% \altaffiliation{}
\affiliation{Graduate School of Information Science and Technology, The University of Tokyo, Tokyo 113-8656, Japan}
\affiliation{Next Generation Artificial Intelligence Research Center, The University of Tokyo, Tokyo 113-8656, Japan}

\date{\today}

\begin{abstract}
    A continuous one-dimensional map with period three includes all periods.
    This raises the following question: Can we obtain any types of periodic orbits solely by learning three data points?
    In this paper, we report the answer to be yes.
    Considering a random neural network in its thermodynamic limit,
    we first show that almost all learned periods are unstable, and each network has its own characteristic attractors (which can even be
    untrained ones).
    The latently acquired dynamics, which are unstable within the trained network, serve as a foundation for the diversity of characteristic attractors and may even lead to the emergence of attractors of all periods after learning.
    When the neural network interpolation is quadratic, a universal post-learning bifurcation scenario appears, which is consistent with a topological conjugacy between the trained network and the classical logistic map.
    In addition to universality, we explore specific properties of certain networks, including the singular behavior of the scale of weight at the infinite limit, the finite-size effects, and the symmetry in learning period three.
\end{abstract}

%\keywords{}

\maketitle

%\tableofcontents

\section{\label{sec:introduction}Introduction}
    With the advent of reservoir computing (RC)
    \cite{nakajima2021reservoir},
    which exploits the dynamics of high-dimensional dynamical systems for learning,
    many interesting properties of random networks have been discovered from a dynamical systems perspective.
    For example, with a fixed-weight random recurrent neural network, also referred to as the echo-state network (ESN)
    \cite{jaeger2001echo}---which functions as a reservoir---we can create an autonomous system that emulates target chaotic systems by simply fitting its readout layer
    \cite{pathak2017using,lu2018attractor}.
    Such an attractor-embedding ability is linked to the existence of the generalized synchronization between an input dynamical system and a reservoir
    \cite{lu2018attractor,hart2020embedding,grigoryeva2021chaos,grigoryeva2023learning}.
    In addition, recent studies have revealed that
    RC can simultaneously embed multiple attractors
    \cite{flynn2021multifunctionality}
    and may have untrained attractors that are not part of the training data
    \cite{
        flynn2021multifunctionality,flynn2023seeing,
        kong2021machine,fan2021anticipating,kim2021teaching,patel2021using, fan2022learning, rohm2021model,
        kabayama2025designing}.
    Surprisingly, a successful reservoir computer only needs 
    a few pairs of bifurcation parameter values and corresponding trajectories 
    to reconstruct the entire bifurcation structure of a target system 
    \cite{
        kong2021machine,fan2021anticipating,kim2021teaching,patel2021using, fan2022learning}.
    These properties are valuable, for example, in the context of robot locomotion control using dynamical system attractors that can significantly reduce the training data
    \cite{steingrube2010self,ijspeert2007swimming,akashi2023embedding}.
    Thus, the powerful generalization and multifunctionality aspects of RC are related to the dynamical systems properties of a learning machine.

    In one-dimensional discrete dynamical systems,
    there are two significant theorems on periodic orbits
    \cite{burns2011sharkovsky,blokh2022sharkovsky,li1975period}:

    \begin{thm}{\rm (Sharkovsky)}
        \label{thm:Sharkovsky}
        If a continuous map $f\mathbin{:} I \rightarrow I$ has a periodic point of period $m$,
        and $m \succ n$, then $f$ also has a periodic point of period $n$.
    \end{thm}
    \begin{thm}{\rm (Li--Yorke)}
        \label{thm:Li--Yorke}
        If a continuous map $f\mathbin{:} I \rightarrow I$ has a point $a \in I$
        for which the points $b = f(a)$, $c = f(f(a))$, and $d = f(f(f(a)))$ satisfy
        \begin{equation}
            \label{eq:Li--Yorke}
            d \leq a < b < c \quad ({\rm or} \quad d \geq a > b > c ),
        \end{equation}
        then 
        $f$ has a periodic point of period $k$ for every $k\in\mathbb{N}$.
    \end{thm}
    Note that the interval $I$ does not need to be closed or bounded.
    Here, the ordering of positive integers $\succ$ in Theorem~\ref{thm:Sharkovsky}
    is called the Sharkovsky ordering and is given below:
    \begin{equation}
        \label{eq:Sharkovsky_ordering}
        \begin{aligned}
            3 &\succ 5 \succ 7 \succ 9 \succ \cdots \succ
            2 \cdot 3 \succ 2 \cdot 5 \succ 2 \cdot 7 \succ \cdots
            \\
            &\succ
            2^2 \cdot 3 \succ 2^2 \cdot 5 \succ 2^2 \cdot 7 \succ \cdots
            \succ
            2^3 \succ 2^2 \succ 2 \succ 1.
        \end{aligned}
    \end{equation}
    We write $m \succ n$ whenever $m$ is to the left of $n$ in Eq.~\eqref{eq:Sharkovsky_ordering}.
    As a consequence of both theorems, a continuous one-dimensional map with period three has all periods.

    Now, the following natural question arises:
    Can we obtain all the periods in the network through training only period three?
    If we successfully train period three in one-dimensional dynamics, then the straightforward answer is ``yes.''
    However, this question is somewhat naive,
    since the above theorems do not reveal whether the obtained periods are stable.
    Instead, we should ask the following question:
    Which kind of stable orbits (attractors) can we obtain by learning period three (LP3)?
    This paper is devoted to theoretically answering this question in terms of all aspects.

    \subsection{\label{subsec:overview}Overview of approach}
        To clarify what LP3 implies, we need to create a one-dimensional recurrent neural network with target period three.
        LP3 with ESN, as in the standard RC scheme, will violate the assumption of the above two theorems due to the high-dimensionality of the resulting dynamics.
        Thus, here, we consider training
        a random feedforward neural network to learn period three.
        We mainly focus on the simplest case: LP3 with a one-layer random feedforward neural network $f_N$ in the readout-only training
        (see Sec.~\ref{subsec:network})
        \cite{
            tokunaga1994reconstructing,itoh2017reconstructing,itoh2020reconstructing, hara2022learning};
        however, our results are applicable to a wide variety of network architectures for which a corresponding kernel $\Theta(x,y)$ is defined
        (see Appendix~\ref{appendix:variations})
        \cite{jacot2018neural,lee2019wide,arora2019exact}.
        With the trained network $f_N^*$, we study the dynamical system $x_{n+1} = f_N^*(x_n)$, which is created by closing a loop that connects its input and output.
        The attractors of $f_N^*$ should depend on the network structure, including the realizations of internal weights $W_i^{\text{in}}$ and $b^{\text{in}}$ and the choice of nonlinearity $\phi$, which makes the above question non-trivial.
        Hereafter, we consider the following specific activation functions: bounded and smooth $\phi=\text{erf},\sin,\cos$, and unbounded and non-smooth $\phi=\text{ReLU}$.
        Variations in activation correspond to differences in the realization of this network in the physical world, such as in optoelectronic systems ($\phi = \sin, \cos$) \cite{nakajima2022physical} and spintronic systems ($\phi = \text{erf}, \text{ReLU}$) \cite{raimondo2021reliability}.

    \subsection{\label{subsec:summary}Summary of major results and organization of the paper}
        \begin{figure}
            \includegraphics[width=\columnwidth]{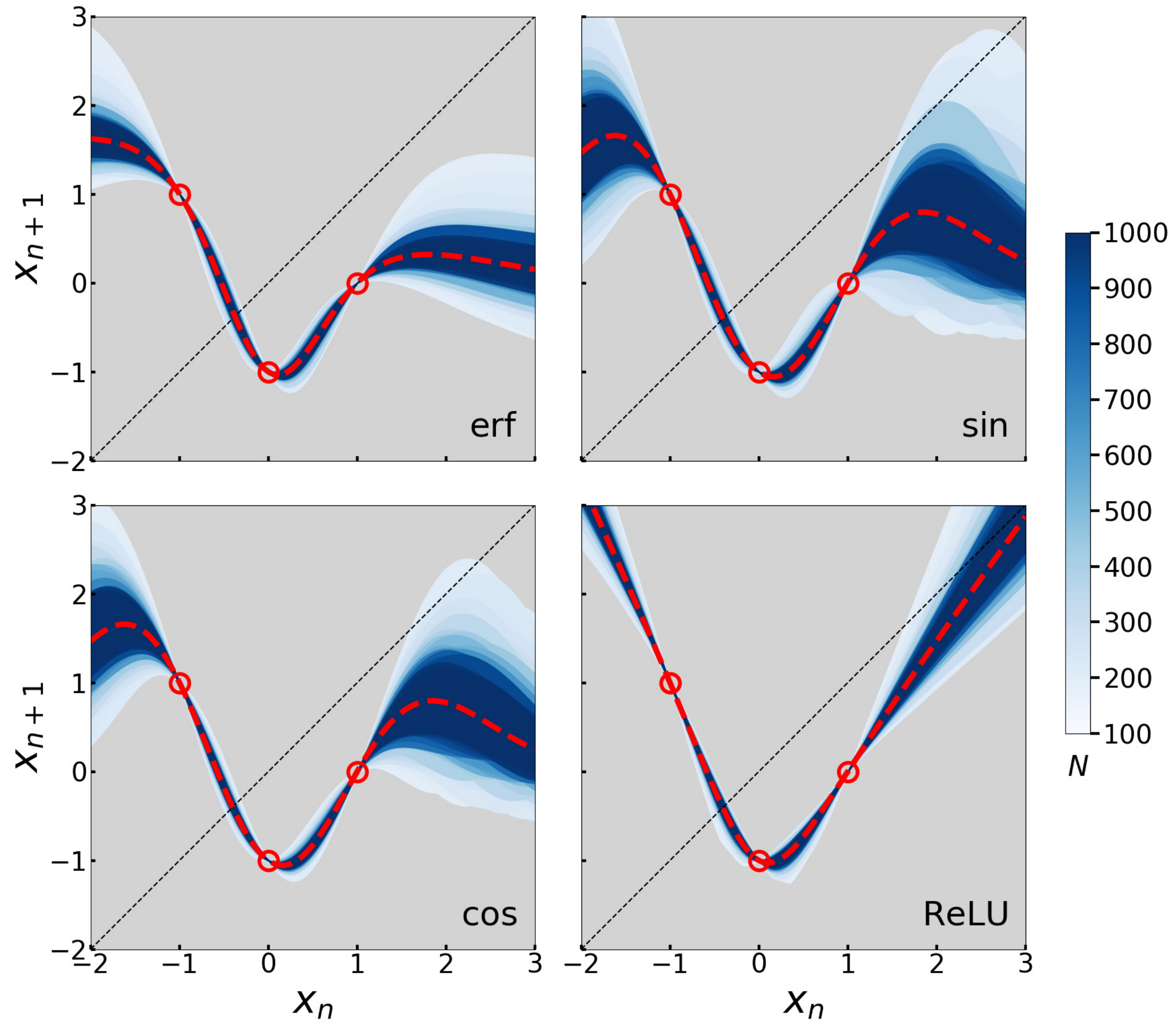}
            \caption{
                Trained maps $f_N^*$ for activation functions $\phi=\text{erf}$ (top left), $\sin$ (top right), $\cos$ (bottom left), and $\text{ReLU}$ (bottom right), with target period three ${\cal D}=\{-1,1,0\}$ and scale of weights $\sigma_w=\sigma_b=\sigma=1.0$.
                The blue-colored areas indicate the maximum--minimum regions of $f_N^*$ for 100 different realizations.
                The red circles and the red dotted lines show ${\cal D}$ and the thermodynamic limit $f_\infty^*$, respectively.
                The shade of blue corresponds to the number of nodes $N$,
                thus indicating that $f_N^*$ degenerates into $f_\infty^*$ as $N$ increases.
            }
            \label{fig:trained_map}
        \end{figure}

        Our findings are threefold. First, we show that the trained map $f_N^*$ degenerates into its thermodynamic limit $f_{\infty}^*$ as $N$ increases (Fig.~\ref{fig:trained_map}) under certain assumptions.
        This insight enables us to explore the invariant properties of the dynamics of trained networks using $f_{\infty}^*$.
        Second, in LP3, we reveal that almost all learned periods are unstable; it has characteristic attractors corresponding to the choice of target period three, nonlinearity $\phi$, and the scale of weights $\sigma_w$ and $\sigma_b$.
        Each characteristic attractor is related through bifurcation, together forming a bifurcation of embeddable attractors for each learning and network configuration.
        Third,
        once the networks learn period three, they can generate a wide variety of attractors through post-learning bifurcation.
        We propose a theory explaining its mechanism, which indicates that under certain conditions, all the latently acquired periods can be externalized by controlling the feedback strength of the trained networks after learning.
        We will also discuss how Sharkovsky ordering appears in the context of externalization. 

        In Sec.~\ref{sec:theoretical}, we provide the theoretical setup for LP3 within the thermodynamic limit of neural networks.
        Next, we present theoretical and numerical results for the dynamics of the trained networks through LP3 in Sec.~\ref{sec:rslt}.
        Finally, in Sec.~\ref{sec:dscs}, we conclude with the implications of our results and discuss the range of applicability of our proposed approach to physical learning machines.

\section{\label{sec:theoretical}Theoretical Setup}
    \subsection{\label{subsec:network}Basic network architecture}
        A one-layer random feedforward neural network is defined by
        \begin{equation}
            \label{eq:NN}
            \begin{gathered}
                f_N(x)
                \equiv
                \frac{1}{\sqrt{N}}\sum_{i=1}^{N} W_i^{\text{out}} \phi\left(h_i(x)\right)
                ,
                \\
                h_i(x)
                \equiv
                \sigma_w W_i^{\text{in}}x + \sigma_b b_{i}^{\text{in}},
            \end{gathered}
        \end{equation}
        where $x\in\mathbb{R}$ is an input of the network;
        $W^{\text{in}}\in\mathbb{R}^{N\times 1}$ and $b^{\text{in}}\in\mathbb{R}^{N}$ are the input weights and biases, respectively, randomly drawn from a normal distribution;
        $\sigma_w$ and $\sigma_b$ are the constants governing the scales of $W^{\text{in}}$ and $b^{\text{in}}$, respectively;
        $\phi\mathbin{:}\mathbb{R}\rightarrow\mathbb{R}$ is an activation function;
        and $W^{\text{out}}\in\mathbb{R}^{1 \times N}$ is the output weights matrix optimized by a learning method described subsequently.
        We set $\sigma_w = \sigma_b = \sigma$ as the scale of weights, except for Sec.~\ref{subsubsec:small}.
        We denote the trained network output by $f_N^*(x)$. 
        This model can be regarded as a special case of an ESN within the limit in which the spectral radius of the adjacency matrix goes to zero.
        Several modifications to this model, such as adding an output bias, changing internal weight distribution, and increasing the number of layers, are also discussed in Appendix~\ref{appendix:variations}.

    \subsection{\label{subsec:thermo}Thermodynamic limit of the trained networks}
        We denote the training dataset and its size by ${\cal D} \subseteq \mathbb{R} \times \mathbb{R}$ and $|{\cal D}|$, respectively, and assume $N\geq|{\cal D}|$.
        We use ${\cal X}$ and ${\cal Y}$ vectors to denote the input and output data and define them as 
        $
            {\cal X}
            \equiv
            [x_1,\ldots,x_{|{\cal D}|}],
            {\cal Y}
            \equiv
            [y_1,\ldots,y_{|{\cal D}|}]
            \in
            \mathbb{R}^{|{\cal D}|}
        $,
        where $(x_i,y_i)\in{\cal D}$.
        In LP3 ($a \mapsto b \mapsto c \mapsto a \mapsto \cdots $), the target input--output pairs are ${\cal D} = \left\{(a,b),(b,c),(c,a)\right\}$, ${\cal X}=[a,b,c]$, and ${\cal Y}=[b,c,a]$.
        For the sake of simplicity, we write ${\cal D}=\{a,b,c\}$ and assume $a < b$.
        Note that a period-three orbit is of two types: $\{a,b,c\}$ and $\{b,a,c\}$.
        Generally, a period-$n$ orbit has $(n-1)!$ types.
        For a given ${\cal D}$, we optimize $W^{\text{out}}$ by
        least square regression with a minimum norm solution (``extreme learning machine'' \cite{huang2006extreme,itoh2017reconstructing, itoh2020reconstructing}) or ``ridgeless'' regression
        \cite{
            saunders1998ridge,suykens2001nonlinear,liang2020just,hastie2022surprises}:
        \begin{equation}
            \label{eq:fN_ridgeless}
            f_N^*(x)
            =
            \lim_{\lambda \searrow 0}
            \hat{\Theta}(x,{\cal X})
            \left(
                \hat{\Theta} + \lambda I_{|{\cal D}|}
            \right)^{-1}
            {\cal Y},
        \end{equation}
        where $\hat{\Theta}(x,{\cal X}) \in \mathbb{R}^{1\times|{\cal D}|}$ and $\hat{\Theta} \equiv \hat{\Theta}({\cal X},{\cal X}) \in \mathbb{R}^{|{\cal D}|\times|{\cal D}|}$
        are the matrices given in the following manner:
        \begin{gather}
            \label{eq:Gram}
            \left.
                \hat{\Theta}(x,{\cal X})
            \right._{i}
            \equiv
            \hat{\Theta} \left(x,x_i\right),
            \left.
                \hat{\Theta}
            \right._{ij}
            \equiv
            \hat{\Theta} \left(x_i,x_j\right),
            \\
            \label{eq:TK}
            \hat{\Theta}(x,y)
            \equiv
            {\cal R}(x)^\top {\cal R}(y)
            =
            \frac{1}{N}\sum_{i=1}^{N}
            \phi\left(h_i(x)\right)\phi\left(h_i(y)\right),
            \\
            \label{eq:reservoir_state}
            {\cal R}(x)_i
            \equiv
            \frac{1}{\sqrt{N}}
            \phi\left(h_i(x)\right)
            ,\quad
            {\cal R}(x) \in \mathbb{R}^{N}.
        \end{gather}
        We note that Eq.~\eqref{eq:fN_ridgeless} is equivalent to fit $W^{\text{out}}$ with the pseudoinverse of the matrix of hidden states ${\cal R}({\cal X})\equiv[{\cal R}(x_1) \cdots {\cal R}(x_{|{\cal D}|})] \in \mathbb{R}^{N \times |{\cal D}|}$ \cite{huang2006extreme}:
        $\left(W^{\text{out}}\right)^{*} = {\cal Y}^\top{\cal R}({\cal X})^{+}$.
        If the matrix $\hat{\Theta}$ has full rank, Eq.~\eqref{eq:fN_ridgeless} has the following closed-form expression:
        \begin{equation}
            \label{eq:fN}
            f_N^*(x)
            =
            \hat{\Theta}(x,{\cal X})
            \hat{\Theta}^{-1}
            {\cal Y}.
        \end{equation}
        By the law of large numbers, $\hat{\Theta}(x,y)$ (Eq.~\eqref{eq:TK}) converges in probability to $\Theta(x,y)$---that is, the expectation over random variables $[\omega,\beta] \sim {\cal N}(\bm{0}, \text{diag}(\sigma_w^2,\sigma_b^2))$---within the limit $N\rightarrow\infty$ because the components of $W^{\text{in}}$ and $b^{\text{in}}$ follow an iid Gaussian distribution
        \cite{jacot2018neural}:
        \begin{equation}
            \label{eq:NTK}
            \begin{aligned}
                &\Theta(x,y)
                =
                \mathbb{E}
                \left[
                    \phi(\omega x+\beta)\phi(\omega y+\beta)
                \right]
                \\
                &
                =
                \frac{1}{2\pi\sigma_w\sigma_b}
                    \int
                    d\omega d\beta \,
                    \phi(\omega x + \beta)
                    \phi(\omega y + \beta)
                    e^{
                        -\frac{1}{2}
                        \left(
                            \frac{\omega^2}{\sigma_w^2}
                            +
                            \frac{\beta^2}{\sigma_b^2}
                        \right)
                    }
                .
            \end{aligned}
        \end{equation}
        In our model, $\Theta(x,y)$ coincides with the neural tangent kernel (NTK) and the neural network Gaussian process (NNGP) kernel \cite{jacot2018neural,lee2019wide}.
        Defining $\Theta(x,{\cal X})$ and $\Theta$ in the same manner as $\hat{\Theta} (x,y)$,
        we acquire $f^*_\infty(x)$, since $f_N^*(x)$ (Eq.~\eqref{eq:fN}) is calculated from only the values of $\hat{\Theta}(x,y)$:
        \begin{equation}
            \label{eq:finf}
            f_\infty^*(x)
            =
            \Theta(x,{\cal X})
            \Theta^{-1}{\cal Y},
        \end{equation}
        where we again assume $\Theta$ to have full rank.
        This assumption is valid if $\phi$ is a non-polynomial continuous function, and ${\cal X}$ consists of $|{\cal D}|$ distinct points
        (see Sec.~\ref{subsec:fullrank})
        .
        With regard to LP3, $f_\infty^*$ is given by
        \begin{equation}
            \label{eq:finf_p3}
            f_\infty^*(x)
            =
            \begin{bmatrix}
                \Theta(x,a)
                \\
                \Theta(x,b)
                \\
                \Theta(x,c)
            \end{bmatrix}^{\top}
            \begin{bmatrix}
                \Theta(a,a) & \Theta(a,b) & \Theta(a,c) \\
                \Theta(b,a) & \Theta(b,b) & \Theta(b,c) \\
                \Theta(c,a) & \Theta(c,b) & \Theta(c,c) 
            \end{bmatrix}^{-1}
            \begin{bmatrix}
                b
                \\
                c
                \\
                a
            \end{bmatrix}.
        \end{equation}

        Note that we can generalize Eq.~\eqref{eq:finf} to the infinite-width deep neural networks in the readout-only training \cite{lee2019wide,arora2019exact} or the lazy full training with small initial output \cite{jacot2018neural,arora2019exact}, simply by introducing the corresponding NNGP kernel or NTK, respectively (see Appendix~\ref{appendix:variations}).
        In addition, the output of the infinite-width ESN also reduces to a form similar to that of Eq.~\eqref{eq:finf} with time-varying recurrent kernels \cite{hermans2012recurrent,dong2020reservoir,dong2022asymptotic}.
    
        For $\phi=\text{erf},\sin,\cos,\text{ReLU}$, there exist the analytic solutions of $\Theta(x,y)$
        \cite{williams1996computing,cho2009kernel,han2022fast,rahimi2007random,louart2018random,pearce2020expressive}:
        \begin{align}
            \label{eq:NTKerf}
            &
            \Theta^{\text{erf}}(x,y)
            \notag
            \\
            &=
            \frac{2}{\pi} \arcsin
            \frac{
                2(\sigma_b^2 + \sigma_w^2 xy)
            }{
                \sqrt{
                    \left[1 + 2(\sigma_b^2 + \sigma_w^2x^2)\right]
                    \left[1 + 2(\sigma_b^2 + \sigma_w^2y^2)\right]
                }
            },
            \\
            \label{eq:NTKsin}
            &
            \Theta^{\sin}(x,y)
            =
            \frac{1}{2}
            \left\{
                e^{-\frac{\sigma_w^2}{2}(x-y)^2}
                -
                e^{-\frac{\sigma_w^2}{2}(x+y)^2-2\sigma_b^2}
            \right\},
            \\
            \label{eq:NTKcos}
            &
            \Theta^{\cos}(x,y)=
            \frac{1}{2}
            \left\{
                e^{-\frac{\sigma_w^2}{2}(x-y)^2}
                +
                e^{-\frac{\sigma_w^2}{2}(x+y)^2-2\sigma_b^2}
            \right\},
            \\
            \label{eq:NTKrelu}
            &
            \begin{aligned}
                \Theta^{\text{relu}}(x,y)
                =
                &
                \frac{1}{2\pi}
                \sqrt{
                    \left(\sigma_b^2 + \sigma_w^2x^2\right)
                    \left(\sigma_b^2 + \sigma_w^2y^2\right)
                }
                \\
                &
                \left\{
                    \sqrt{1-\cos^2\psi}
                    +
                    (\pi - \psi)
                    \cos\psi
                \right\}
                ,
                \\
                \text{where}
                \,\,
                \psi
                &
                \equiv
                \arccos
                \frac{
                    \sigma_b^2+\sigma_w^2xy
                }{
                    \sqrt{\left(\sigma_b^2 + \sigma_w^2x^2\right)\left(\sigma_b^2 + \sigma_w^2y^2\right)}
                }.
            \end{aligned}
        \end{align}
        Figure \ref{fig:trained_map} shows the shape of the trained map for each activation function and how the trained network, with a finite number of nodes, degenerates into its unique thermodynamic limit.
        We note that the trajectories for a bounded activation function---such as $\phi=\text{erf},\sin,\cos$---are also bounded, since $\Theta(x,y)$ (Eq.~\eqref{eq:NTK}) is the expectation of the product of $\phi$.
        In contrast, certain trajectories for $\phi=\text{ReLU}$, whose NTK is unbounded, will head toward infinity.

    \subsection{\label{subsec:dynamical}Dynamical system analysis in the infinite trained networks}
        To investigate the dynamics of $f_\infty^*$,
        we compute the trajectory $\left\{x_{n}\right\}_{n=0}^T$ of $T + 1 \gg 1$ steps with an initial state $x_0$
        and calculate the Lyapunov exponent and period of attractors from the trajectories.
        The Lyapunov exponent expresses the sensitivity of
        a dynamical system to initial conditions and is calculated using the following equation:
        \begin{equation}
            \label{eq:LE}
            \lambda_T
            =
            \frac{1}{T}
            \sum_{n=1}^{T}
            \ln{\left|\frac{df_\infty^*}{dx}(x_n)\right|}.
        \end{equation}
        We regard a trajectory of $f_\infty^*$ as chaotic when $\lambda_T>0$.
        Note that the derivative of $f_\infty^*$ is calculated by
        \begin{equation}
            \label{eq:dxfinf}
            \frac{df_\infty^*}{dx}(x)
            =
            \frac{\partial\Theta}{\partial x}(x,{\cal X})
            \Theta^{-1}{\cal Y}.
        \end{equation}
        The formulas for $\frac{\partial\Theta}{\partial x}(x,y)$ used in Eq.~\eqref{eq:dxfinf} are presented in Appendix~\ref{appendix:formulas}.
        We also calculate the period of attractors as the minimum integer $n \in [1,n_{\text{max}}]$
        that satisfies the following inequality ($n_{\text{max}}=10\text{ or }20$ and $\varepsilon = 10^{-12}$ except as otherwise noted):
        \begin{equation}
            \label{eq:period}
            \begin{gathered}
                \left|(f_\infty^*)^n(x_{T}) - x_{T} \right|
                \leq
                \varepsilon
                \cdot
                \max \left\{\left|x_{T} \right|,\left|(f_\infty^*)^n(x_{T})\right| \right\}
                ,
                \\
                \text{where}
                \quad
                (f_\infty^*)^1 \equiv f_\infty^*
                ,\,
                (f_\infty^*)^{k+1} \equiv f_\infty^* \circ (f_\infty^*)^{k}.
            \end{gathered}
        \end{equation}

    \subsection{\label{subsec:fullrank}Full rankness of the Gram matrix $\Theta$}
        Before presenting our main results, we will first establish the validity of our full-rank assumption for $\Theta$, as discussed in Ref.~\cite{carvalho2024positivity}.
        For this purpose, we assume that the activation function $\phi$ is continuous.
        Note that for the infinitely differentiable non-polynomial $\phi$, the full rankness of the matrix $\hat{\Theta}$ (with finite $N$) is discussed in Ref.~\cite{tamura1997capabilities,huang2006extreme}.
        As $\Theta$ is symmetric, it is necessary and sufficient to show that $\Theta$ is positive definite:
        \begin{equation}
            \label{eq:pd_gram_1}
            \begin{aligned}
                u^\top \Theta u
                &=
                \mathbb{E}
                \left[
                    \sum_{i=1}^{|{\cal D}|}
                    \sum_{j=1}^{|{\cal D}|}
                    u_i \phi(\omega x_i+\beta)
                    \phi(\omega x_j+\beta) u_j
                \right]
                \\
                &=
                \mathbb{E}
                \left[
                    \left(
                        \sum_{i=1}^{|{\cal D}|}
                        u_i \phi(\omega x_i+\beta)
                    \right)^2
                \right]
                > 0
            \end{aligned}
        \end{equation}
        for any non-zero $u \in \mathbb{R}^{|{\cal D}|}\setminus\left\{0\right\}$.
        From Eq.~\eqref{eq:pd_gram_1}, our assumption breaks down if and only if $u \neq 0$ satisfies
        \begin{equation}
            \sum_{i=1}^{|{\cal D}|}
            u_i \phi(\omega x_i+\beta)
            = 0
        \end{equation}
        for almost every $[\omega,\beta] \sim {\cal N}(\bm{0}, \text{diag}(\sigma_w^2,\sigma_b^2))$. Since ${\cal N}(\bm{0}, \text{diag}(\sigma_w^2,\sigma_b^2))$ has full support, and $\phi$ is continuous, it is equivalent to
        \begin{equation}
            \label{eq:pd_gram_2}
            \sum_{i=1}^{|{\cal D}|}
            u_i \phi(W x_i+b)
            = 0,
            \,\,
            \text{for every}
            \,\,
            [W, b] \in \mathbb{R}^2.
        \end{equation}

        With the input data ${\cal X}$ consisting of $|{\cal D}|$ distinct points, the following theorem states that if Eq.~\eqref{eq:pd_gram_2} holds, then $\phi$ is a polynomial function.

        \begin{thm}{\rm (Ref.~\cite{carvalho2024positivity})}
            \label{thm:pd}
            Let $z, w \in \mathbb{R}^{|{\cal D}|}$ be totally non-aligned, meaning that
            \begin{equation}
                \label{eq:nonaligned}
                \begin{vmatrix}
                    z_i & w_i \\
                    z_j & w_j
                \end{vmatrix}
                \neq 0,
                \,\,
                \text{for all}
                \,\,
                i \neq j,
            \end{equation}
            and let $\phi\mathbin{:}\mathbb{R}\rightarrow\mathbb{R}$ be continuous. If there exists $u \in \mathbb{R}^{|{\cal D}|}\setminus\left\{0\right\}$ such that
            \begin{equation}
                \label{eq:pd}
                \sum_{i=1}^{|{\cal D}|}
                u_i\phi(\theta_1 z_i + \theta_2 w_i)
                = 0,
                \,\,
                \text{for every}
                \,\,
                [\theta_1, \theta_2] \in \mathbb{R}^2,
            \end{equation}
            then $\phi$ is a polynomial function.
        \end{thm}
        Therefore, our assumption is valid in learning period $n=1,2,3,\ldots$ as long as the activation $\phi$ is a non-polynomial continuous function.
        Note that the activations we use---specifically $\phi=\text{erf},\sin,\cos,\text{ReLU}$---satisfy this condition, resulting in full-rank Gram matrices $\Theta$.

        In a deep neural network (Eq.~\eqref{eq:deep}), a non-polynomial, continuous, and differentiable-almost-everywhere $\phi$ ensures that the associated kernel is strictly positive definite, leading to a full-rank Gram matrix \cite{carvalho2024positivity}.
        Notably, the differentiability of $\phi$ is unnecessary in the readout-only training.
        This property arises because the kernel's positivity in the first hidden layer propagates to subsequent layers when $\phi$ is non-constant \cite{jacot2018neural,carvalho2024positivity}.

\section{\label{sec:rslt}Results}
    In this section, we will discuss the results obtained through LP3 by dividing them into two sections:
    Sec.~\ref{subsec:general}, which covers the universal properties that are independent of our network model selection, and Sec.~\ref{subsec:special}, which covers the interesting properties obtained from our specific network model.
    \subsection{\label{subsec:general}General properties of LP3}
        \subsubsection{\label{subsubsec:finiteness}Finiteness of the attractors}
            Mathematically, it can be demonstrated that neural networks with specific activation functions have a finite number of possible attractors.
            In particular, we will utilize the following properties of a smooth map from a finite interval $I$ to itself.
            \begin{thm}{\rm (Melo--Strien)}
                \label{thm:melo}
                If $f\mathbin{:} I \rightarrow I$ is a $C^2$ map with non-flat critical points,
                then there exist $\lambda > 1$ and $n_0 \in \mathbb{N}$ such that
                \begin{equation}
                    \label{eq:melo}
                    \left|\frac{d}{dx}f^{n}(p)\right| > \lambda
                \end{equation}
                for every periodic point $p$ of $f$ of period $n \geq n_0$.
            \end{thm}
            Here, we say that $c \in I$ is a critical point if it satisfies $\frac{d}{dx}f(c)=0$, and that a critical point $c$ for a smooth map is non-flat if there exists $k \geq 2$ such that $\frac{d^k}{dx^k}f(c)\neq0$
            \cite{melo1993one,van1988smooth}.
            Note that if $f$ is a non-constant analytic map with critical points, then all the critical points are non-flat, since $f(x)$ has the Taylor expansion for every $x_0 \in I$, and its coefficients for some $k\geq2$ degrees are non-zero \cite{van1988smooth}.
            Hence, we obtain the following corollary of Theorem~\ref{thm:melo}:
            \begin{cor}
                \label{cor:finiteness}
                If $f\mathbin{:} I \rightarrow I$ is a non-constant analytic map with critical points,
                then there exist $\lambda > 1$ and $n_0 \in \mathbb{N}$ such that
                \begin{equation}
                    \left|\frac{d}{dx}f^{n}(p)\right| > \lambda
                \end{equation}
                for every periodic point $p$ of $f$ of period $n \geq n_0$.
            \end{cor}
            For our neural network model, if the kernels $\hat{\Theta}(x,y)$ and $\Theta(x,y)$ are bounded and analytic (e.g., if $\phi=\text{erf},\sin,\cos$), then the network outputs are also bounded and analytic, since $f_N^*(x)$ (Eq.~\eqref{eq:fN}) and $f_\infty^*(x)$ (Eq.~\eqref{eq:finf}) are described by the weighted sum of the kernels.
            Furthermore, in learning period $n \geq 2$, $f_N^*(x)$ and $f_\infty^*(x)$, with full-rank matrices $\hat{\Theta}$ and $\Theta$, respectively, cannot be constant functions, because in that case they will not be able to replicate the target input--output pairs ${\cal D}$.
            In addition, in LP3, according to Rolle's theorem, $f_N^*$ and $f_\infty^*$ have at least one critical point due to their folding around ${\cal D}$; thus, we can restrict the bounded $f_N^*$ and $f_\infty^*$ to some finite interval that contains all the periods and critical points.
            Therefore, with the full-rank Gram matrix and the bounded and analytic kernel, $f_N^*$ and $f_\infty^*$ in LP3 has, at most, finitely many stable periods (attractors) based on Corollary~\ref{cor:finiteness}.

        \subsubsection{\label{subsubsec:emergence}Emergence of the untrained attractors}
            \begin{figure}
                \includegraphics[width=\columnwidth]{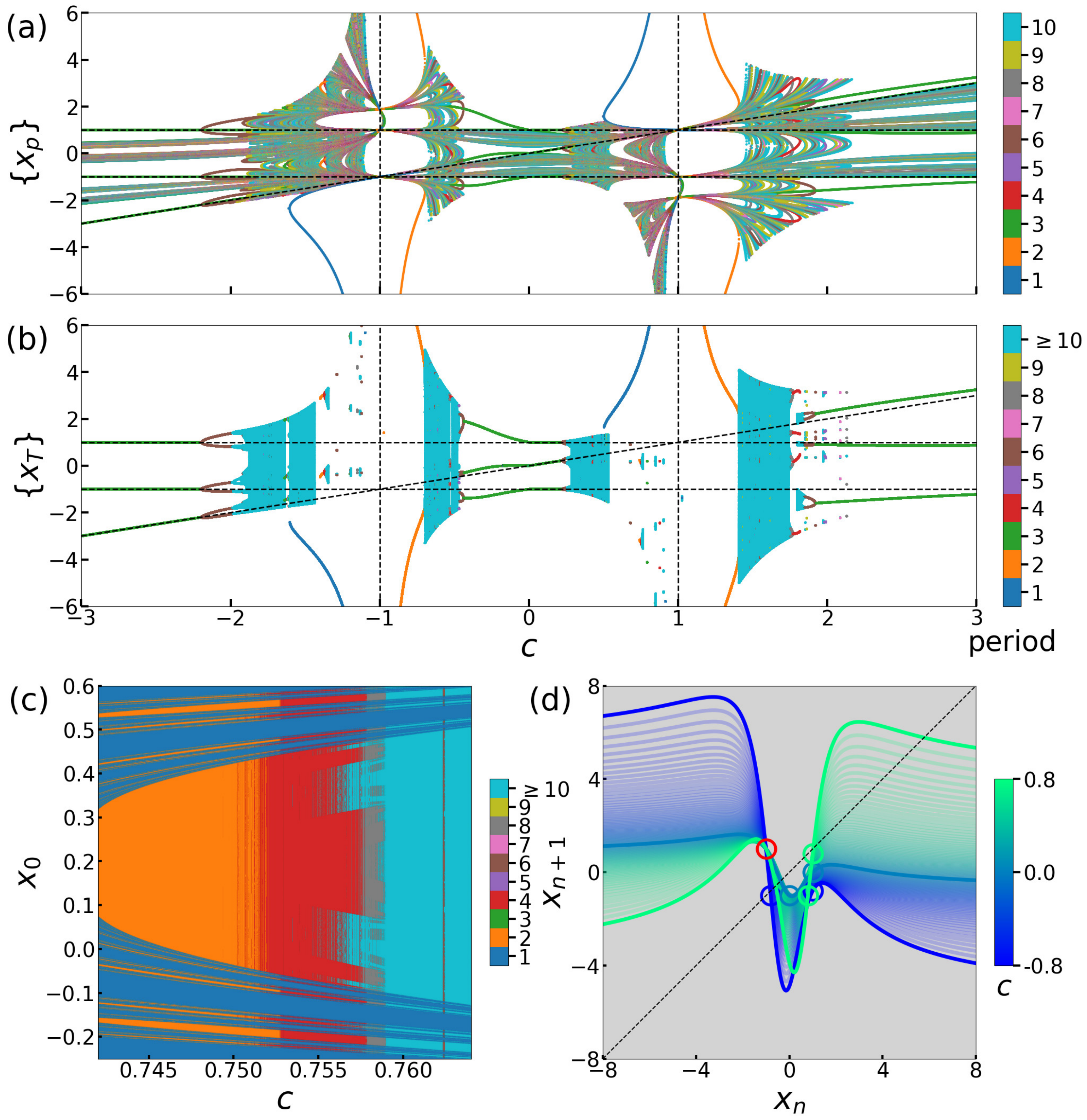}
                \caption{
                Change in dynamical system $f_\infty^*$ for $\phi=\text{erf}$
                with respect to $c$, with $a=-1$, $b=1$, and $\sigma=1.0$.
                The trajectory $\{x_n\}_{n=0}^{T}$ of $T=10^5$ steps is computed with different $x_0$ and $c$ in given intervals, excluding $c=a,b$.
                (a) Changes in the learned periods $\{x_p\}$ of period $p$, calculated by solving $(f_\infty^*)^{p}(x_p) = x_p$
                (see Appendix~\ref{appendix:numerical} for details).
                (b) Bifurcation diagram of the characteristic attractors calculated with $-10\leq x_0 \leq 10$.
                The dotted lines indicate ${\cal D}$; the diagonal lines correspond to varying $c$.
                (c) Change in the basin of attraction, where $f_\infty^*$ has multiple untrained attractors.
                (d) Change in the map $f_\infty^*$ in $-0.8 \leq c \leq 0.8$.
                The circles indicate ${\cal D}$, and the red circle indicates a c-independent point $(a,b)=(-1,1)$.
                As $c$ approaches $a$ or $b$, the folding of $f_\infty^*$ around ${\cal D}$ becomes larger, making the characteristic attractors wider.
                }
                \label{fig:1dbifurcation}
            \end{figure}

            Figure \ref{fig:1dbifurcation} depicts how $f_\infty^*$ for $\phi=\text{erf}$ changes as $c$ varies.
            Note that as long as $\hat{\Theta}$ and $\Theta$ have full rank, trained networks completely learn the target orbit ${\cal D}$, since $f_N^*({\cal X})=f_\infty^*({\cal X})={\cal Y}$.
            In the case of ${\cal D}$ becoming the attractor ($-3 \leq c < -2.2$ and $0 < c \leq 0.22$), which corresponds to the successful attractor embedding by a learning machine, we observe that $f_\infty^*$ has no untrained attractors.
            However, varying $c$ causes the bifurcation of the embeddable attractors of $f_\infty^*$, resulting in the emergence of untrained attractors:
            untrained period-three ($-0.43 \leq c < 0$ and $2 \leq c \leq 3$), chaotic ($c=-0.5,0.3$, etc.), and multiple attractors ($0.742 \leq c \leq 0.764$, etc.).
            Consequently,
            together with Corollary~\ref{cor:finiteness},
            only a handful of attractors appear at a time, and almost all periods latently exist as unstable periods (Fig.~\ref{fig:1dbifurcation}(a),(b)).
            Accordingly, we define a pre-learning bifurcation as the bifurcation of characteristic attractors that occurs due to changes in training or network configurations before the learning process begins.
            The pre-learning bifurcation structure strongly depends on the network settings
            (see Fig.~\ref{fig:2dbifurcation} for the bifurcation diagrams with respect to $c$ and $\sigma$, along with the corresponding $\lambda_T$).
            We note that as the NTKs for $\phi=\text{erf},\sin,\cos,\text{ReLU}$ depend on the scaling and translation of inputs, varying $a$ or $b$, which we fixed, also yields a different bifurcation.

        \subsubsection{\label{subsubsec:bifurcation}Stability changes in unstable periods through a post-learning bifurcation}
            \begin{figure*}
                \includegraphics[width=\textwidth]{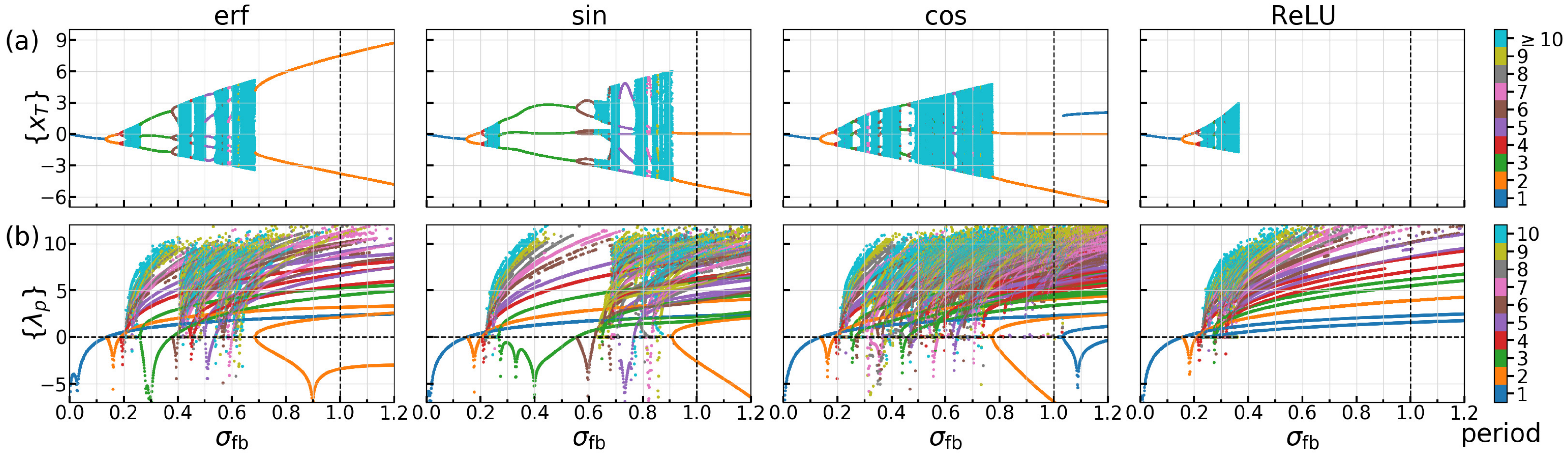}
                \caption{
                    Change in dynamical systems $\sigma_\text{fb}f_\infty^*$ with respect to feedback strength $\sigma_\text{fb}$, with ${\cal D}=\{-1,1,-0.8\}$ and $\sigma=1$.
                    (a) Bifurcation diagrams calculated with $-10\leq x_0 \leq 10$ and $T=10^5$.
                    (b) Stability changes in the learned periods $\{x_p\}$ of period $p$, calculated by $\lambda_p \equiv \ln \left| \frac{d}{dx} (\sigma_\text{fb}f_\infty^*)^{p}(x_p) \right|$ using the solutions of $(\sigma_\text{fb}f_\infty^*)^{p}(x_p) = x_p$.
                    The vertical dotted lines correspond to the network state in LP3 ($\sigma_\text{fb} = 1$).
                    The horizontal dotted line indicates the boundary of the stability $\lambda_p = 0$; $\lambda_p < 0$ and $\lambda_p > 0$ mean (locally) stable and unstable, respectively.
                    In the range $0 \leq \sigma_\text{fb} \leq 1$, the unstable periods at $\sigma_\text{fb} = 1$ will emerge as specific attracting periodic orbits, referred to as ``externalized.''
                    }
                \label{fig:externalization}
            \end{figure*}
    
            So far, we have only focused on the characteristic attractors of the neural networks in LP3, which are just small parts of latently existing, infinitely many periodic orbits.
            To illuminate the meaning of the latently existing unstable periods, we further extend our learning procedure to varying the readout weights of the trained neural networks \cite{hara2022learning}.
            Figure \ref{fig:externalization} depicts how their attractors change as the scale of the readout weights---the feedback strength $\sigma_\text{fb}$---varies.
            We set the network state of LP3 to $\sigma_\text{fb} = 1$; this dynamical system is expressed as $\sigma_\text{fb}f_\infty^*$.
            Since the network is prohibited from having any periodic orbits, except a fixed point $x=0$ at $\sigma_\text{fb} = 0$, all the learned periods $\{x_p\}$ must disappear at some $\sigma_\text{fb} \in (0,1)$ with $\frac{d}{dx} (\sigma_\text{fb} f_\infty^*)^{p}(x_p) = 1$ due to the differentiability of $f_\infty^*$; some of them may become attractors before dying out.
            In numerical experiments, we observed a decline in the Lyapunov exponents for each unstable periodic orbit, defined as $\lambda_p \equiv \ln \left| \frac{d}{dx} (\sigma_\text{fb}f_\infty^*)^{p}(x_p) \right|$ (Fig.~\ref{fig:externalization}(b)).
            These avalanches in the stability of unstable periods result in the emergence of various attractors, indicated by Lyapunov exponents falling below $\lambda_p = 0$.
            We refer to this process as ``externalization,'' as it externalizes the latently acquired dynamics within the trained neural network through a post-learning bifurcation.
            Here again, the choice of ${\cal D}$ and the network structure leads to diverse externalization, as illustrated in Fig.~\ref{fig:externalization}.

        \subsubsection{\label{subsubsec:mechanism}Mechanism of externalization}
            We have introduced the concept of externalization that transforms acquired unstable periodic orbits into attractors after learning.
            Still, it is unclear whether or not all the periodic orbits are allowed to emerge as attractors through this process.
            In the following, we prove that in general, for every period $p \in \mathbb{N}$, there exists at least one period-$p$ orbit that will continuously change into an attractor through externalization, provided that the bifurcation causing the disappearance of periodic orbits requires at least one stable periodic orbit.

            We denote by $f^* \mathbin{:} \mathbb{R} \rightarrow \mathbb{R}$ a differentiable map with a period-three orbit, which represents the trained network using LP3.
            Let $x_p(\sigma_\text{fb})$ be a periodic point of period $p$ for the one-parameter family $\sigma_\text{fb} f^*$ around $\sigma_\text{fb} = 1$.
            If it satisfies
            \begin{equation}
                \frac{d}{dx} (f^*)^{p}(x_p(1)) \neq 1
            \end{equation}
            at $\sigma_\text{fb}=1$, then $x_p(\sigma_\text{fb})$ is a differentiable function of $\sigma_\text{fb}$ in the neighborhood of $\sigma_\text{fb} = 1$, according to the implicit function theorem \cite{guckenheimer1979bifurcation}.
            Thus, LP3 satisfying this condition in general induces an infinite number of differentiable curves $\{x_p(\sigma_\text{fb})\}_{p\in\mathbb{N}}$ in the post-learning bifurcation.

            For $p > 1$, these curves must vanish at some $\sigma_p \in (0,1)$ with $\frac{d}{dx} (\sigma_p f_\infty^*)^{p}(x_p(\sigma_p)) = 1$.
            Otherwise, a periodic point of period $p > 1$ would exist at $\sigma_{\text{fb}} = 0$, leading to a contradiction.
            If periodic points vanish through a bifurcation, such as a tangent or period-doubling bifurcation, which requires at least one stable periodic orbit (attractor) just before disappearing, then there exists at least one curve $x_p(\sigma_\text{fb})$ that links the attractor at $\sigma_\text{fb} = \sigma_p$ to the latently acquired periodic orbit of $p$ at $\sigma_\text{fb} = 1$.
            Otherwise, all periodic points of period $p > 1$ would remain unstable until they vanish, again leading to a contradiction.
            For $p = 1$, there are two scenarios for $x_1(\sigma_\text{fb})$: It either disappears at some $\sigma_1 \in (0,1)$ or changes into a fixed point at the origin at $\sigma_{\text{fb}} = 0$.
            The former scenario is included in the previously mentioned case, and the latter is considered a continuous transition into the attractor at $\sigma_{\text{fb}} = 0$.
            Note that not all unstable periodic orbits at $\sigma_\text{fb} = 1$ necessarily become attractors; for instance, a periodic point that disappears through a tangent bifurcation may remain unstable until it vanishes.
            Therefore, we obtain the following:

            \begin{thm}
                \label{thm:externalization}
                Let $f^*\mathbin{:} \mathbb{R} \rightarrow \mathbb{R}$ be a differentiable map with the following properties:
                \begin{description}
                    \item[EX1] $f^*$ possesses a period-three orbit.
                    \item[EX2] The disappearance of a periodic point in the one-parameter family $\sigma_{\rm fb} f^*$ occurs through a bifurcation that requires at least one stable periodic orbit.
                \end{description}
                Denote by $P_{(n,\sigma_{\rm fb})}$ the set of periodic points of period $n$ for $\sigma_{\rm fb} f^*$.
                Suppose that for every $n \in \mathbb{N}$ and $x_n \in P_{(n,1)}$, the following genericity condition holds:
                \begin{equation}
                    \label{eq:genericity}
                    \frac{d}{dx} (f^*)^{n}(x_n) \neq 1.
                \end{equation}
                Then, for any $p \in \mathbb{N}$, there exist a constant $\sigma_p \in [0,1)$ and a differentiable function $x_p(\cdot): [\sigma_p,1] \rightarrow \mathbb{R}$ such that
                \begin{enumerate}
                    \item $x_p(\sigma_{\rm fb}) \in P_{(p,\sigma_{\rm fb})}$ for any $\sigma_{\rm fb} \in [\sigma_p,1]$.
                    \item There exists $\delta > 0$ such that $x_p(\sigma_p+\delta)$ is attracting:
                    \begin{equation}
                        \label{eq:attracting}
                        \left|
                            \frac{d}{dx} [(\sigma_p+\delta)f^*]^{p}(x_p(\sigma_p+\delta))
                        \right| < 1.
                    \end{equation}
                \end{enumerate}
            \end{thm}

            We propose that the abstract features of the curves $\{x_p(\sigma_{\text{fb}})\}$ promote the diversity in externalization, which corresponds to the network structure, as illustrated in Fig.~\ref{fig:externalization}.
            Notably, there are alternative methods for constructing an $f^*$ that satisfies property \textbf{EX1}.
            One such method is to train learning machines with a general dataset (${\cal X}=[a,b,c]$ and ${\cal Y}=[b,c,d]$) that satisfies Li--Yorke's condition (Eq.~\eqref{eq:Li--Yorke}).
            We consider any method of constructing $f^*$ with a period-three orbit to be LP3 in a general sense.
            We also maintain that it remains unclear whether property \textbf{EX2} universally applies to learning machines.
            As a straightforward example, we will examine the detailed properties in externalization when $f^*$ is quadratic.

        \subsubsection{\label{subsubsec:quadratic}Externalization using quadratic interpolation}

            \begin{figure}
                \includegraphics[width=\columnwidth]{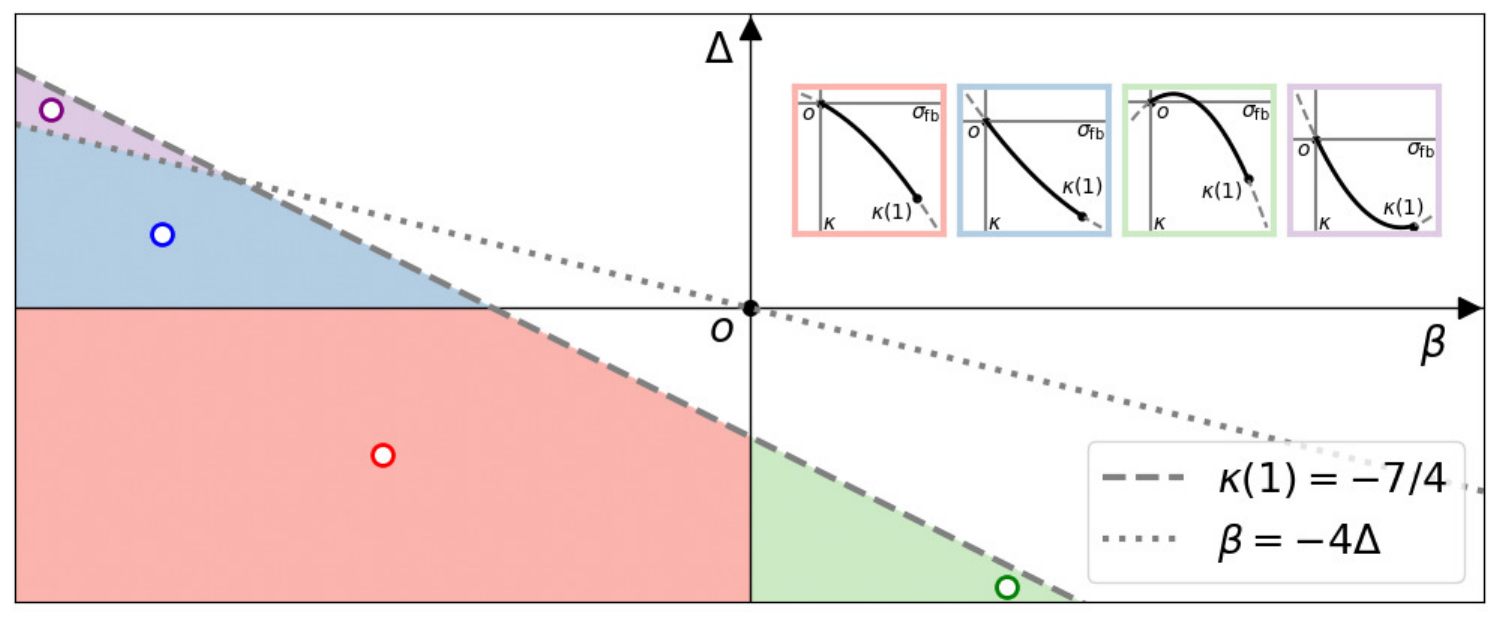}
                \caption{
                    Four possible types of mapping $\kappa(\sigma_\text{fb})$ (Eq.~\eqref{eq:kappa}) restricted to the interval $[0,1]$, depicted as colored regions in the $\beta\Delta$-plane.
                    The insets provide examples for each type of $\kappa(\sigma_\text{fb})$ in the $\sigma_{\text{fb}}\kappa$-plane, corresponding to the colored points in the $\beta\Delta$-plane.
                    The black solid lines in the insets indicate $\kappa([0,1])$ in the $\sigma_{\text{fb}}\kappa$-plane.
                    Note that $\kappa(1) = \Delta + \frac{\beta}{2} \leq -\frac{7}{4}$, due to the presence of period three \cite{shi2007chaos}, $\kappa(-\frac{\beta}{4\Delta})=-\frac{\beta^2}{16\Delta}$ (an extremum), and $\kappa(0) = 0$.
                    }
                \label{fig:kappa}
            \end{figure}
    
            Let us consider a simple construction of $f^*$ using quadratic interpolation:
            \begin{equation}
                \label{eq:quadratic}
                \begin{gathered}
                    f^*(x)
                    =
                    g(x)
                    \equiv
                    \alpha + \beta x + \gamma x^2,
                    \\
                    \text{where}
                    \quad
                    \begin{bmatrix}
                        \alpha
                        \\
                        \beta
                        \\
                        \gamma
                    \end{bmatrix}
                    \equiv
                    \begin{bmatrix}
                        1 & a & a^2 \\
                        1 & b & b^2 \\
                        1 & c & c^2
                    \end{bmatrix}^{-1}
                    \begin{bmatrix}
                        b
                        \\
                        c
                        \\
                        a
                    \end{bmatrix}.
                \end{gathered}
            \end{equation}
            Alternatively, consider a quadratic interpolation of a general dataset (${\cal X}=[a,b,c]$ and ${\cal Y}=[b,c,d]$) that satisfies Li--Yorke's condition (Eq.~\eqref{eq:Li--Yorke}):
            \begin{equation}
                \label{eq:quadratic_LiYorke}
                \begin{gathered}
                    g(x)
                    \equiv
                    \alpha + \beta x + \gamma x^2,
                    \\
                    \text{where}
                    \quad
                    \begin{bmatrix}
                        \alpha
                        \\
                        \beta
                        \\
                        \gamma
                    \end{bmatrix}
                    \equiv
                    \begin{bmatrix}
                        1 & a & a^2 \\
                        1 & b & b^2 \\
                        1 & c & c^2
                    \end{bmatrix}^{-1}
                    \begin{bmatrix}
                        b
                        \\
                        c
                        \\
                        d
                    \end{bmatrix},
                    \\
                    d \leq a < b < c \quad ({\rm or} \quad d \geq a > b > c ).
                \end{gathered}
            \end{equation}
            Changing the feedback strength $\sigma_\text{fb}$ after LP3 corresponds to a scaling change in each coefficient:
            $\sigma_\text{fb}g(x) = \sigma_\text{fb}\alpha + \sigma_\text{fb}\beta x + \sigma_\text{fb}\gamma x^2$.
            Regardless of the dataset used, any bifurcations induced by $\sigma_\text{fb}$ can be simplified to a bifurcation of a map $q_{\kappa}(x) \equiv x^2 + \kappa$ with respect to $\kappa$, due to the topological conjugacy in the quadratic family \cite{peitgen2004}.
            Specifically, the following equation holds:
            \begin{equation}
                \label{eq:quad_topological_conjugacy}
                \sigma_\text{fb}g = h^{-1}_{\sigma_\text{fb}} \circ q_{\kappa} \circ h_{\sigma_\text{fb}},
                \,
                \text{where}
                \,\,
                h_{\sigma_\text{fb}}(x)
                \equiv
                \sigma_\text{fb}
                \left(
                    \gamma x + \frac{\beta}{2}
                \right).
            \end{equation}
            Here, the parameter $\kappa$ is also a function of $\sigma_\text{fb}$ as follows:
            \begin{equation}
                \label{eq:kappa}
                \begin{aligned}
                    \kappa(\sigma_\text{fb})
                    =
                    \sigma_\text{fb}
                    \left(
                        \Delta \sigma_\text{fb}
                        +
                        \frac{\beta}{2}
                    \right),
                    \,
                    \text{where}
                    \,\,
                    \Delta \equiv \alpha \gamma - \frac{\beta^2}{4}.
                \end{aligned}
            \end{equation}
            If the conjecture regarding $q_{\kappa}$ in Ref.~\cite{guckenheimer1979bifurcation}---stating that its bifurcations are either tangent or period-doubling---is correct, then quadratic interpolations (Eqs.~\eqref{eq:quadratic}--\eqref{eq:quadratic_LiYorke}) will externalize all the periods, as per Theorem~\ref{thm:externalization}.
            
            Considering the presence of period three at $\sigma_\text{fb}=1$,
            Eq.~\eqref{eq:kappa} defines four possible types of relationships between $\sigma_\text{fb}$ and $\kappa$ (see Fig.~\ref{fig:kappa}).
            In any case, the image $\kappa([0,1])$ includes the interval $[-\frac{7}{4},0]$, where $q_{\kappa}(x)$ is also topologically conjugated to the logistic map $y_{n+1} = \mu y_n(1 - y_n)$ with $\mu(\kappa) = \sqrt{-4\kappa + 1} + 1$.
            With the Sharkovsky ordering of the birth of periods in the logistic map \cite{pastor1997harmonic,sharkovsky2012difference}, we obtain the following Sharkovsky ordering in the externalization:
            \begin{equation}
                \label{eq:Sharkovsky_externalization}
                \begin{gathered}
                    1
                    \geq
                    \sigma_\text{fb}[3]
                    \geq
                    \sigma_\text{fb}[5]
                    \geq \cdots \geq
                    \sigma_\text{fb}[2 \cdot 3]
                    \geq
                    \sigma_\text{fb}[2 \cdot 5]
                    \geq \cdots
                    \\
                    \geq
                    \sigma_\text{fb}[2^2 \cdot 3]
                    \geq \cdots
                    \geq
                    \sigma_\text{fb}[2^2]
                    \geq
                    \sigma_\text{fb}[2]
                    \geq
                    \sigma_\text{fb}[1]
                    \geq
                    0,
                \end{gathered}
            \end{equation}
            where $\sigma_\text{fb}[n]$ denotes the least value of $\sigma_\text{fb}$ for which $\sigma_\text{fb}f^*$ possesses period $n$ as its attractor.
            We note that the relationship between Theorem~\ref{thm:externalization} and Sharkovsky ordering in externalization (Eq.~\eqref{eq:Sharkovsky_externalization}) remains an open problem.

        \subsubsection{\label{subsubsec:small}Small $\sigma_w$ induces quadratic-like interpolation}

            \begin{figure*}
                \includegraphics[width=\textwidth]{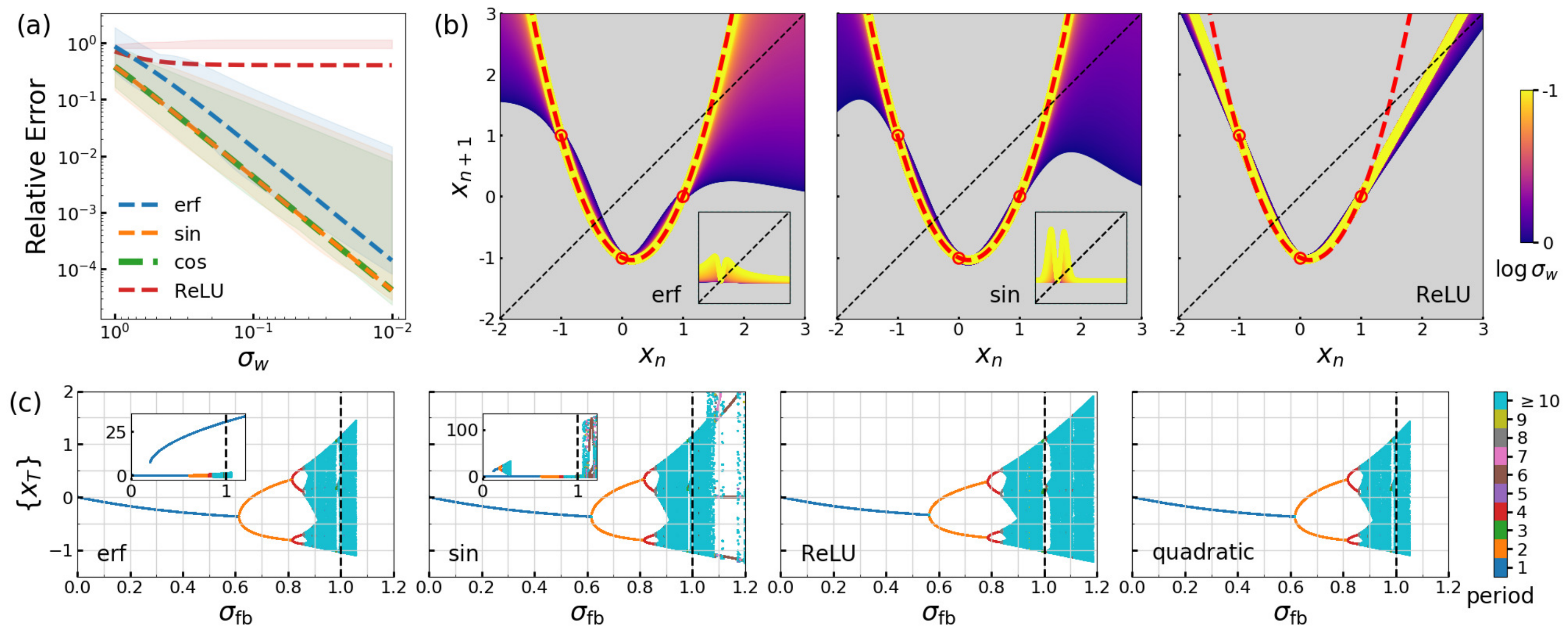}
                \caption{
                    Comparison of neural network interpolations (Eqs.~\eqref{eq:fN} and \eqref{eq:finf}) with quadratic interpolation (Eq.~\eqref{eq:quadratic}) using $\sigma_b=1.0$ and ${\cal D}=\{-1,1,0\}$.
                    (a) Relative errors $e$ (Eq.~\eqref{eq:error}) assessing the closeness of neural network interpolation to the quadratic interpolation.
                    Colored areas indicate the maximum--minimum regions of the relative errors for 500 different realizations of $f_{N=100}^*$.
                    Dotted lines correspond to their thermodynamic limit $N\rightarrow\infty$.
                    (b) Change in the maps $f_\infty^*(x)$ with respect to $\sigma_w$, with $\phi=\text{erf}$ (left), $\sin$ (middle), and $\text{ReLU}$ (right).
                    The red circles and the red dotted lines show ${\cal D}$ and the corresponding quadratic map, respectively.
                    The insets are zoom-outs in the range $[-50, 150]$.
                    (c) Externalization $\sigma_\text{fb}f_\infty^*(x)$ with $\sigma_w=10^{-1}$, calculated with $-10\leq x_0 \leq 10$, $T=10^4$, and $\varepsilon=10^{-6}$.
                    With a small $\sigma_w$ and analytic $\phi$, the externalization shows the quantitative universality characterized by quadratic interpolation (see rightmost panel).
                    However, it also generates attractors not explained by this universality, as shown in the zoomed-out bifurcation diagrams (insets) and in the region in which the attractors of the quadratic map vanishes due to a boundary crisis ($\sigma_\text{fb} \approx 1.05$) \cite{grebogi1983crises}.
                    The trained network with $\phi=\text{ReLU}$ does not converge to quadratic interpolation as $\sigma_w$ decreases, but it still exhibits externalization qualitatively similar to quadratic interpolation.
                    }
                \label{fig:confirm}
            \end{figure*}

            To conclude this section, we will demonstrate that under certain conditions, specifically when the scaling of $W^{\text{in}}$ is sufficiently small ($\sigma_w \ll 1$), the interpolation of neural networks in LP3 can be quadratic.
            We assume that the trained network output $f^*_N(x)$ is analytic and completely learns the target orbit ${\cal D}$: $f_N^*({\cal X}) ={\cal Y}$.
            In the thermodynamic limit, the latter is consistent with the condition of $\Theta$ having full rank, as discussed in Sec.~\ref{subsec:fullrank}.
            Using a Taylor expansion with respect to $\sigma_w$, we derive the following second-order approximation for a sufficiently small $\sigma_w$ \cite{tadokoro2024trans}:
            \begin{equation}
                \label{eq:approx}
                f^*_N(x)
                \approx
                f^*_N(0) + \frac{df^*_N}{dx}(0) x + \frac{1}{2}\frac{d^2f^*_N}{dx^2}(0) x^2
            \end{equation}
            If this network reproduces target period three, the second-order approximate function (Eq.~\eqref{eq:approx}) should converge to the corresponding quadratic interpolation (Eq.~\eqref{eq:quadratic}).
            We will numerically verify this property by evaluating the relative error between the coefficients of Eq.~\eqref{eq:quadratic} and Eq.~\eqref{eq:approx}: 
            \begin{equation}
                \label{eq:error}
                e
                \equiv
                \frac{
                    \left\|
                        \begin{bmatrix}
                            f^*_N(0)
                            \\
                            \frac{df^*_N}{dx}(0)
                            \\
                            \frac{1}{2}\frac{d^2f^*_N}{dx^2}(0)
                        \end{bmatrix}
                        -
                        \begin{bmatrix}
                            1 & a & a^2 \\
                            1 & b & b^2 \\
                            1 & c & c^2
                        \end{bmatrix}^{-1}
                        \begin{bmatrix}
                            b
                            \\
                            c
                            \\
                            a
                        \end{bmatrix}
                    \right\|
                }
                {
                    \left\|
                        \begin{bmatrix}
                            1 & a & a^2 \\
                            1 & b & b^2 \\
                            1 & c & c^2
                        \end{bmatrix}^{-1}
                        \begin{bmatrix}
                            b
                            \\
                            c
                            \\
                            a
                        \end{bmatrix}
                    \right\|
                }
                ,
            \end{equation}
            where $\| \cdot \|$ denotes the vector norm.
            Note that we calculate the error (Eq~\eqref{eq:error}) for the thermodynamic limit by substituting $f_N^*$ with $f_\infty^*$; however, the rigorous validity of this substitution is uncertain (refer to Appendix~\ref{appendix:formulas} for useful formulas).

            Figure~\ref{fig:confirm} presents a comparison between neural network interpolations (Eqs.~\eqref{eq:fN} and \eqref{eq:finf}) and quadratic interpolation, analyzed from both a functional perspective (Fig.~\ref{fig:confirm}(a),(b)) and a dynamical systems perspective (Fig.~\ref{fig:confirm}(c)).
            As previously discussed, the trained network with analytic activations ($\phi=\text{erf},\sin,\cos$) approximates a quadratic form around target period three as $\sigma_w$ decreases.
            Due to the approximate nature of these relationships, deviations of $f_\infty^*$ from quadratic interpolations are observed in the zoomed-out return maps.
            Accordingly, the externalization with small $\sigma_w$ shows both similarities and differences compared to quadratic interpolation.
            Specifically, attractors not described by topological conjugacy to the quadratic map may emerge either far from the target-period-three region or after the boundary crisis \cite{grebogi1983crises} of the quadratic map.
            Conversely, decreasing $\sigma_w$ does not bring the trained network with non-analytic activation ($\phi=\text{ReLU}$) closer to the quadratic map, yet it still exhibits externalization similar to that of quadratic interpolation.
            We currently lack a clear explanation for this phenomenon and will address it in future research.
            Nevertheless, these properties could be useful for designing the bifurcation structure of a physical neural network (PNN), wherein the emergence of attractors at a specific feedback strength $\sigma_\text{fb}$ is predictable in advance.

    \subsection{\label{subsec:special}Special properties in particular networks}

        \begin{figure*}
            \includegraphics[width=\textwidth]{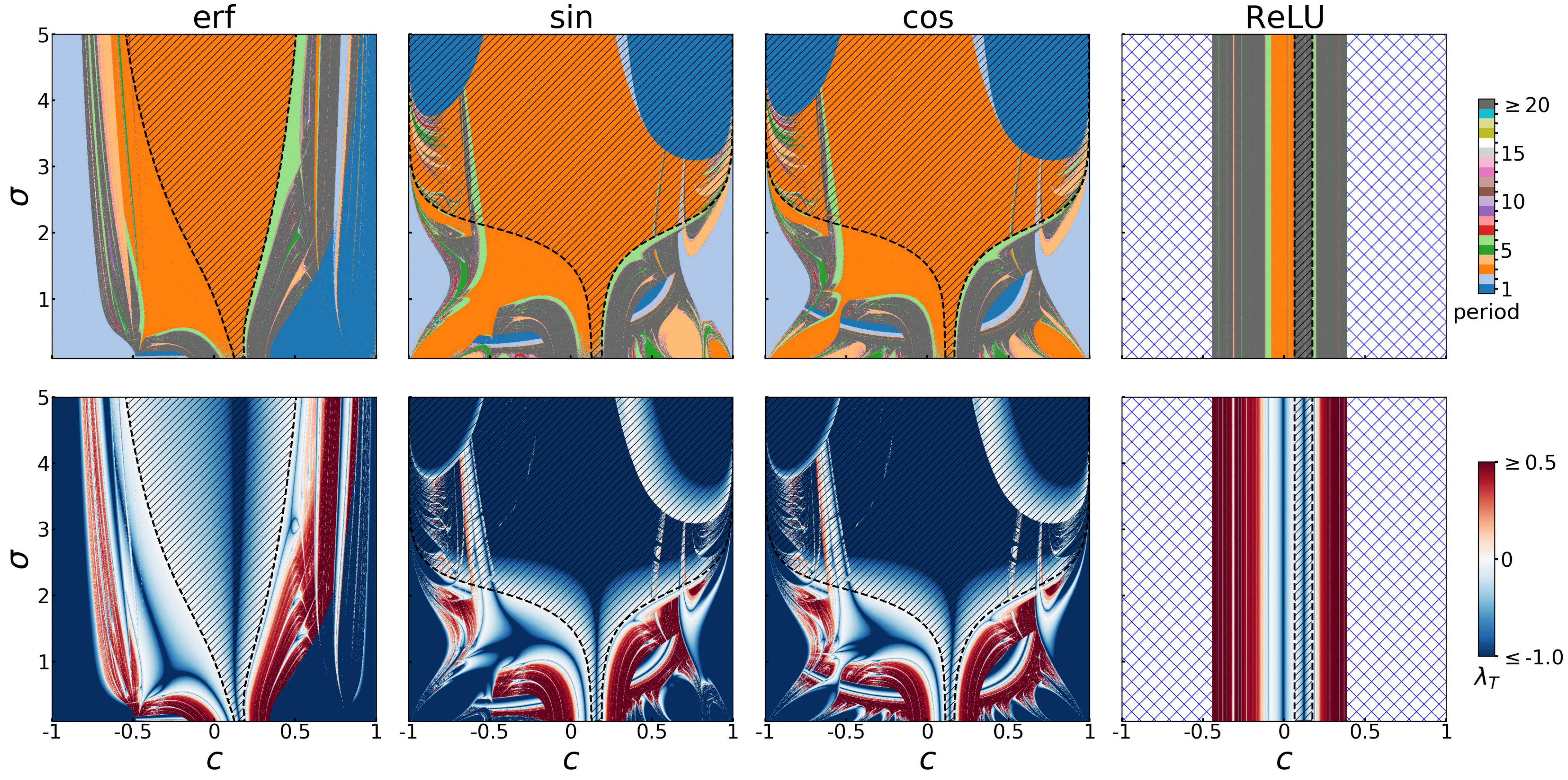}
            \caption{
                Two-dimensional slices of the pre-learning bifurcation of dynamical systems $f_\infty^*$ with respect to $c$ and $\sigma$, with $a=-1$, $b=1$, $x_0=0$, and $T = 10^4$.
                Period of attractors (top row).
                Lyapunov exponents (bottom row).
                The black-hatched areas indicate the regions in which $\left|\frac{df_\infty^*}{dx}(a) \frac{df_\infty^*}{dx}(b) \frac{df_\infty^*}{dx}(c)\right| < 1$ holds.
                They correspond to the regions of ${\cal D}=\{a,b,c\}$ being locally stable, thus implying that for $\phi=\text{erf},\sin,\cos$, increasing $\sigma$ tends to stabilize ${\cal D}$.
                The blue-hatched area indicates the region $(c,\sigma)$, in which the trajectory starting from $x_0=0$ heads toward infinity.
            }
            \label{fig:2dbifurcation}
        \end{figure*}

        \subsubsection{\label{subsubsec:singular}Singular behavior within the limit $\sigma \rightarrow \infty$ and finite-size effects}
        
            \begin{figure*}
                \includegraphics[width=\textwidth]{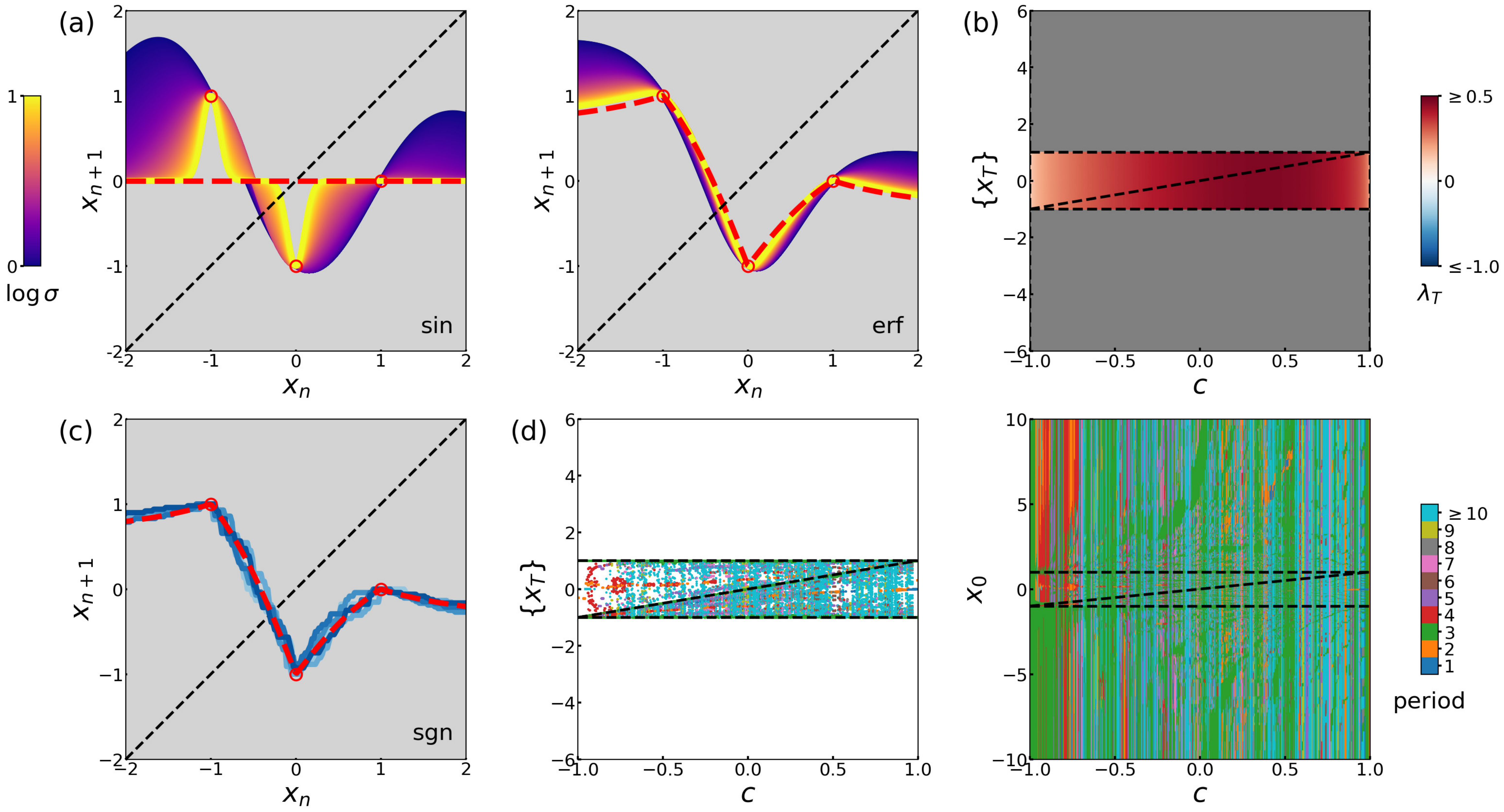}
                \caption{
                    Dynamical systems $f_\infty^*$ within the limit $\sigma\rightarrow\infty$, with $a=-1$ and $b=1$.
                    (a) Change in the maps $f_\infty^*$ for $\phi=\sin$ (left) and $\text{erf}$ (right) with respect to $\sigma$, with $c=0$ fixed.
                    The red circles and the red dotted lines show the target period three and $\lim_{\sigma\rightarrow\infty}f_\infty^*$, respectively.
                    (b) Pre-learning bifurcation of $\lim_{\sigma\rightarrow\infty}f_\infty^*$ for $\phi=\text{erf}$
                    with respect to $c$, calculated with $-10\leq x_0 \leq 10$.
                    In this limit, $f_\infty^*$ for $\phi=\text{erf}$ becomes piecewise-monotonic and piecewise-smooth,
                    with target period three being its singular points. In this setting, $f_\infty^*$ exhibits the candidate of robust chaos.
                    (c) Trained maps $f_N^*$ and $f_\infty^*$, with $\phi=\text{sgn}$, $c=0$, and $N=100$.
                    The red circles and the red dotted line show target period three and $f_\infty^*$, respectively. The blue solid lines indicate five different realizations of $f_N^*$.
                    (d) Pre-learning bifurcation diagram of the attractors (left) and change in the basin of attraction (right) calculated with $\phi=\text{sgn}$, $N=10^3$, $-10\leq x_0 \leq 10$, $T=10^3$, and fixed realizations of the input layer.
                    The complex structure in (d) corresponds to the discontinuity of $f_N^*$, and the trained network with $\phi=\text{sgn}$ qualitatively changes within its thermodynamic limit (b).
                }
                \label{fig:inf_sigma}
            \end{figure*}

            \begin{figure*}
                \includegraphics[width=\textwidth]{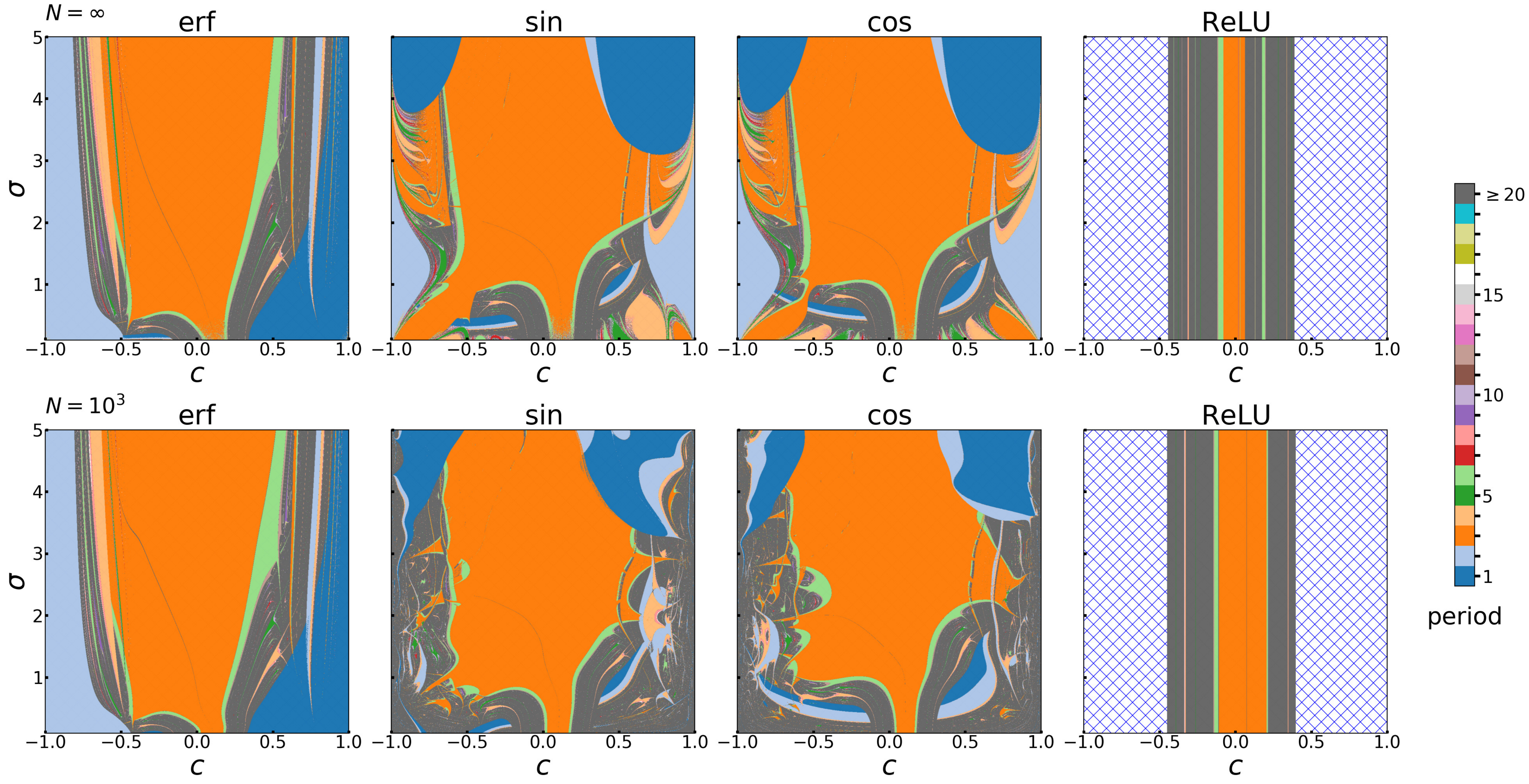}
                \caption{
                    Two-dimensional slices of the pre-learning bifurcation of $f_\infty^*$ (top row) and $f_N^*$ (bottom row) with respect to $c$ and $\sigma$, with $a=-1$, $b=1$, $N=10^3$, $x_0=0$, and $T = 10^4$.
                    Realizations of the input layer of $f_N^*$ are fixed for comparison.
                    The blue-hatched area indicates the region $(c,\sigma)$, in which the trajectory starting from $x_0=0$ heads toward infinity.
                }
                \label{fig:finite_bf}
            \end{figure*}

            \begin{figure*}
                \includegraphics[width=\textwidth]{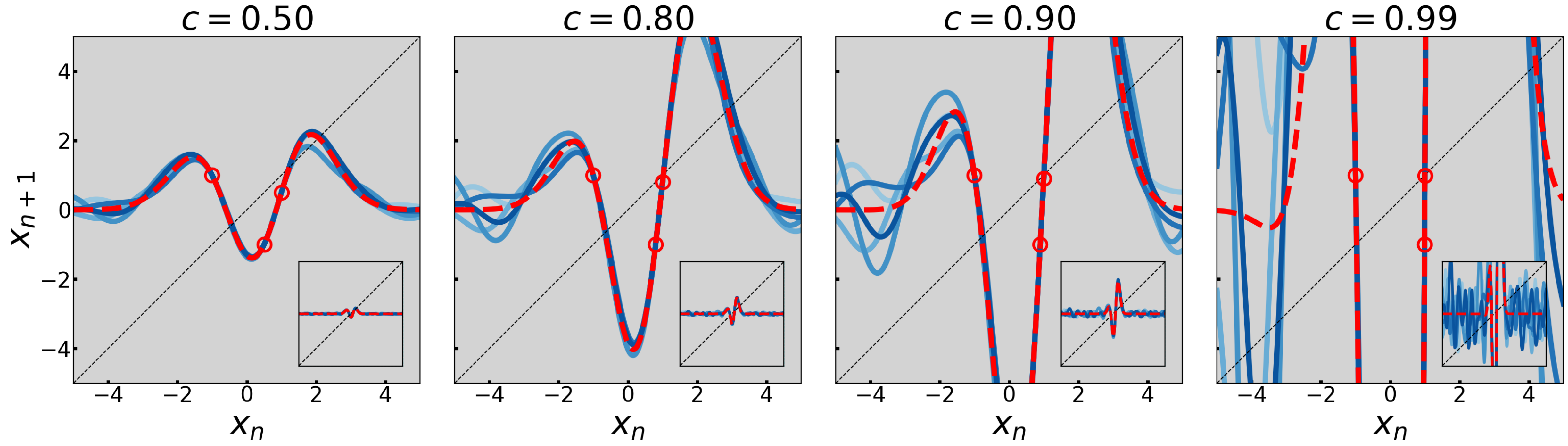}
                \caption{
                    Trained maps $f_N^*$ and $f_\infty^*$ for $\phi=\sin$ with $a=-1$, $b=1$, $N=10^3$, and $\sigma=1.0$.
                    The red circles and the red dotted lines show target period three and $f_\infty^*$, respectively.
                    The insets are zoom-outs in the range $[-20, 20]$.
                    The blue solid lines indicate five different realizations of $f_N^*$;
                    the wavy deviations of $f_N^*$ from $f_\infty^*$ increase as $c$ approaches $a$ or $b$, leading to the finite-size effects in $f_N^*$ for $\phi=\sin,\cos$ (see Fig.~\ref{fig:finite_bf}).
                }
                \label{fig:finite_map}
            \end{figure*}

            The differences in network structures lead to a surprising diversity in characteristic attractors, as illustrated in Fig.~\ref{fig:2dbifurcation}.
            For $\phi=\text{ReLU}$, the $\sigma$-dependence of NTK (Eq.\eqref{eq:NTKrelu}) is canceled out, resulting in the uniform bifurcation structure along the $\sigma$-direction. 
            In contrast, varying $\sigma$ dramatically changes the characteristic attractors for $\phi=\text{erf},\sin,\cos$; the target orbit ${\cal D}$ tends to be locally stable as $\sigma$ increases (the black-hatched areas in Fig.~\ref{fig:2dbifurcation}) because the derivative of NTK at data point $y$ approaches zero for a large $\sigma$:
            \begin{gather}
                \label{eq:dxNTKerf_ondata}
                \frac{\partial\Theta^{\text{erf}}}{\partial x}(y,y)
                =
                \frac{4\sigma^2y}{\pi\left[1+2\sigma^2(1+y^2)\right]\sqrt{1+4\sigma^2(1 + y^2)}}
                ,
                \\
                \label{eq:dxNTKsincos_ondata}
                \frac{\partial\Theta^{\sin}}{\partial x}(y,y)
                =
                -
                \frac{\partial\Theta^{\cos}}{\partial x}(y,y)
                =
                \sigma^2 y e^{-2\sigma^2(1+y^2)}.
            \end{gather}

            However, if the value of $\sigma$ is too large, $\Theta(x,y)$ and, therefore, the dynamical system $f_\infty^*$ qualitatively change (Fig.~\ref{fig:inf_sigma}).
            For $\phi=\sin,\cos$, the trained maps become the Kronecker delta-like discontinuous functions that behave as constant functions except at ${\cal D}$, meaning that perturbations of $x_0$ from ${\cal D}$ lead to their different attractors for a large $\sigma$:
            \begin{gather}
                \label{eq:NTK_sincos_sigmainf}
                \begin{gathered}
                    \lim_{\sigma\rightarrow\infty}
                    \Theta^{\sin}(x,y)
                    =
                    \lim_{\sigma\rightarrow\infty}
                    \Theta^{\cos}(x,y)
                    =
                    \frac{1}{2}
                    \bm{1}_{y}(x)
                    ,
                    \\
                    \text{where}
                    \,\,
                    \bm{1}_{y}(x)
                    \equiv
                    \begin{cases}
                        1 & \left(x = y\right) \\
                        0 & \left(\text{otherwise}\right)
                    \end{cases}
                    ,
                \end{gathered}
                \\
                \label{eq:finf_sincos_sigmainf_period3}
                \begin{aligned}
                    \lim_{\sigma\rightarrow\infty}
                    f_\infty^{*,\sin}(x)
                    &=
                    \frac{1}{2}
                    \begin{bmatrix}
                        \bm{1}_{a}(x) & \bm{1}_{b}(x) & \bm{1}_{c}(x)
                    \end{bmatrix}
                    (2I)
                    \begin{bmatrix}
                        b
                        \\
                        c
                        \\
                        a
                    \end{bmatrix}
                    \\
                    &=
                    b\bm{1}_{a}(x) + c\bm{1}_{b}(x) + a\bm{1}_{c}(x)
                    ,
                \end{aligned}
                \\
                \begin{gathered}
                    \lim_{\sigma\rightarrow\infty}
                    f_\infty^{*,\sin}({\cal X})
                    =
                    \lim_{\sigma\rightarrow\infty}
                    f_\infty^{*,\cos}({\cal X})
                    =
                    {\cal Y},
                    \\
                    \lim_{\sigma\rightarrow\infty}
                    f_\infty^{*,\sin}(x)
                    =
                    \lim_{\sigma\rightarrow\infty}
                    f_\infty^{*,\cos}(x)
                    =
                    0
                    \,
                    \left(x \notin \{a,b,c\}\right).
                \end{gathered}
            \end{gather}

            For $\phi=\text{erf}$, the trained map becomes the piecewise-monotonic and piecewise-smooth function, with ${\cal D}$ being its singular points:
            \begin{gather}
                \label{eq:NTKerf_sigmainf}
                \lim_{\sigma \rightarrow \infty} \Theta^{\text{erf}}(x,y)
                =
                \frac{2}{\pi} \arcsin
                \frac{1+xy}{\sqrt{\left(1+x^2\right)\left(1+y^2\right)}}
                ,
                \\
                \label{eq:dx_NTKerf_sigmainf}
                \begin{aligned}
                    \frac{\partial}{\partial x}
                    \lim_{\sigma\rightarrow\infty}
                    \Theta^{\text{erf}}(x,y)
                    &=
                    \frac{-2}{\pi(1+x^2)}
                    \cdot
                    \frac{x-y}{|x-y|}
                    \\
                    &=
                    \begin{cases}
                        \frac{2}{\pi} \cdot \frac{1}{1+x^2} & \left(x < y\right) \\
                        -\frac{2}{\pi} \cdot \frac{1}{1+x^2} & \left(x > y\right)
                    \end{cases},
                \end{aligned}
            \end{gather}
            which enables the candidates of robust chaos \cite{banerjee1998robust,banerjee2000bifurcations} to appear (Fig.~\ref{fig:inf_sigma}(b)).
            In particular, NTK for $\phi=\text{erf}$ is equivalent to that for binary activation ($\phi=\text{sgn}$) \cite{louart2018random,dong2022asymptotic} within the limit $\sigma\rightarrow\infty$:
            \begin{equation}
                \label{eq:NTKsgn}
                \lim_{\sigma \rightarrow \infty} \Theta^{\text{erf}}(x,y)
                =
                \frac{2}{\pi} \arcsin\frac{1+xy}{\sqrt{\left(1+x^2\right)\left(1+y^2\right)}}
                .
            \end{equation}

            Although $f_N^*$ for $\phi=\text{sgn}$ is beyond the scope of Sharkovsky's theorem (Theorem~\ref{thm:Sharkovsky}) and Li--Yorke's theorem (Theorem~\ref{thm:Li--Yorke}), it asymptotically approaches a continuous map $f_\infty^*$, thereby exhibiting multiple stable periodic orbits for a large $N$ (Fig.~\ref{fig:inf_sigma}(c),(d)).
            Finite-size effects also emerge for $\phi=\sin,\cos$, which are caused by the wavy deviations of $f_N^*$ from $f_\infty^*$ (Figs.~\ref{fig:finite_bf} and \ref{fig:finite_map} ).

        \subsubsection{\label{subsubsec:symmetries}Symmetries in LP3}
            Below, we assume that the kernel $\Theta(x,y)$ remains unchanged when the signs of two slots are switched (i.e., $\Theta(x,y) = \Theta(-x,-y)$).
            Note that any kernel in our model (Eq.~\eqref{eq:NTK}) satisfy this condition, due to the symmetry in the distribution of $W^{\text{in}}$ (also see Eqs.~\eqref{eq:NTKerf}--\eqref{eq:NTKrelu}):
            \begin{equation}
                \label{eq:NTK_symmetry}
                \begin{aligned}
                    \Theta(-x,-y)
                    &=
                    \mathbb{E}_{\omega,\beta}
                    \left[
                        \phi(-\omega x+\beta)
                        \phi(-\omega y+\beta)
                    \right]
                    \\
                    &=
                    \mathbb{E}_{-\omega,\beta}
                    \left[
                        \phi(\omega x+\beta)
                        \phi(\omega y+\beta)
                    \right]
                    =
                    \Theta(x,y).
                \end{aligned}
            \end{equation}
            Then, considering a specific choice of target period three ($a=-b$), we observe a qualitative correspondence between the outside $(c < a, b < c)$ and the inside $(a < c < b)$ of the pre-learning bifurcation with respect to $c$, as illustrated in Fig.~\ref{fig:1dbifurcation}(a),(b).
            This phenomenon arises from the symmetry in the trained networks (as detailed in Theorem~\ref{prop:sym_tc}) and the qualitative similarity between the two types of a period-three orbit, ${\cal D}=\{a,b,c\}$ and ${\cal D}=\{b,a,c\}$, particularly when $c = a \pm \varepsilon$ or $c = b \pm \varepsilon$ with $\varepsilon \ll 1$:
            \begin{equation}
                \label{eq:sym_p3}
                \begin{gathered}
                    \{a, b, a \pm \varepsilon\}
                    \approx
                    \{b, a, a \mp \varepsilon\}
                    \approx
                    \{a, a, b\},
                    \\
                    \{a, b, b \pm \varepsilon\}
                    \approx
                    \{b, a, b \mp \varepsilon\}
                    \approx
                    \{a, b, b\},
                \end{gathered}
            \end{equation}
            where the signs of $\varepsilon$ are determined to preserve the positional relationships in the return maps.
            
            \begin{thm}
                \label{prop:sym_tc}
                Let $\Theta(x,y)$ be the kernel satisfying $\Theta(x,y) = \Theta(-x,-y)$.
                Then, the two networks, each trained differently on specific period-three orbits: ${\cal D} = \{a,-a,c\}$ and ${\cal D} = \{-a,a,-c\}$, are topologically conjugate in the following manner:
                \begin{equation}
                    \label{eq:sym_tc}
                    \left. f_\infty^*(x) \right|_{\{a,-a,c\}}
                    =
                    -
                    \left. f_\infty^*(-x) \right|_{\{-a,a,-c\}}.
                \end{equation}
            \end{thm}
            The proof of Theorem~\ref{prop:sym_tc} can be found in Appendix~\ref{appendix:proof}.
            Combining these two properties, we derive the following approximations for sufficiently small $\varepsilon \ll 1$:
            \begin{equation}
                \label{eq:inversion}
                \begin{aligned}
                    \left.
                        f_\infty^*(x)
                    \right|_{\{a,-a,a\pm\varepsilon\}}
                    &\overset{\mathrm{Eq.~\eqref{eq:sym_p3}}}{\approx}
                    \left.
                        f_\infty^*(x)
                    \right|_{\{-a,a,a\mp\varepsilon\}}
                    \\
                    &\overset{\mathrm{Eq.~\eqref{eq:sym_tc}}}{\approx}
                    -
                    \left.
                        f_\infty^*(-x)
                    \right|_{\{a,-a,-a\pm\varepsilon\}}.
                \end{aligned}
            \end{equation}
            Equation~\eqref{eq:inversion} finalizes the qualitative correspondence between the outside $(c = a - \varepsilon, c = - a + \varepsilon)$ and the inside $(c = - a - \varepsilon, c = a + \varepsilon)$ of the pre-learning bifurcation structure (see Fig.~\ref{fig:1dbifurcation}(a),(b)).

\section{\label{sec:dscs}Discussion}

    Although LP3 is not a necessary condition for achieving period three (learning period two or even random neural networks \cite{ishihara2005magic} may have period three; see Fig.~\ref{fig:period_n_distribution} and Fig.~\ref{fig:randomNN} in Appendix~\ref{appendix:numerical}), it provides a sufficient condition for embedding attractors with all periods as a post-learning bifurcation, along with externally controllable parameters ${\cal D}$ and $\sigma_\text{fb}$---which is our answer to the very first question.
    LP3 also provides new yet important perspectives on the learning of dynamics.
    First, even if neural networks completely learn a target orbit ${\cal D}$, they may fail to replicate ${\cal D}$, since its stability depends on the local structure of the trained map $f_\infty^*$ around ${\cal D}$.
    Second, LP3 is not a goal, but rather a groundwork or a primer for updating the connectivity of random networks to generate all types of periodic orbits, including chaos, after learning.
    Generating network dynamics with a minimal dataset is not just efficient but also compatible with the theoretical analysis, since $f_\infty^*(x)$ is described by the multiplication of $\Theta(x,{\cal X})$ and $\Theta^{-1}$; learning dynamics from time series lead to a large $|{\cal D}|$ and a nearly singular $\Theta$, making the analysis of $f_\infty^*$ ineffective.
    Thermodynamic limit analysis enables us to examine the invariant properties of trained random networks.
    Specifically, it offers a universal framework for comparing network dynamics by compressing network characteristics into kernel properties.

    Generic learning machines satisfying conditions \textbf{EX1} and \textbf{EX2} will externalize attractors of all periods after learning, regardless of whether they are within the thermodynamic limit.
    However, several unresolved issues remain regarding the properties of externalization.
    For instance, it is unclear how one would construct or test learning machines that meet condition \textbf{EX2}, aside from selecting those that interpolate quadratically, so that only a tangent or period-doubling bifurcation seems to occur.
    Additionally, it is worth exploring whether a universal ordering exists for the emergence of attractors in general externalizations, as observed in the externalization using quadratic interpolation (Eq.~\eqref{eq:Sharkovsky_externalization}).
    Nonetheless, we believe our work will encourage further analysis of the relationship between the bifurcation-embedding capability of learning machines and their physical characteristics.

    As an engineering application, we may replace the feedforward network part with a physical system or neuromorphic device \cite{ortin2015unified,nakajima2020physical,marcucci2020theory}.
    In such cases, LP3 would highlight the distinct characteristics of each PNN as its pre- and post-learning bifurcations.
    Although we demonstrated LP3 using only a layered network structure, LP3 itself is not restricted by network architecture as long as the resulting network dynamics remain one-dimensional.
    For example, we can exploit the nonlinear current--voltage characteristics of the ${\text{Mn}_{12}}$ molecular redox array \cite{hirano2012mn12,kan2021simple} as a physical activation function or as a high-dimensional nonlinear mapping by utilizing the diversity of its threshold voltage in a design-less manner \cite{bose2015evolution}.

    Additionally, in the context of embedding bifurcation into physical systems \cite{akashi2023embedding}, our study tells us how to embed a bifurcation having all periods within PNNs.
    It is furthermore possible to specify the initial condition and the bifurcation parameter $\sigma_\text{fb}$ for generating almost desired periodic orbits of PNNs by inducing quadratic-like interpolation (Sec.~\ref{subsubsec:small}), because the basin of attraction of a quadratic map $q_{\kappa}(x) = x^2 + \kappa$ is quite simple---it has at most one stable periodic orbit \cite{guckenheimer1979bifurcation}---and the bifurcation parameter values for stable periodic orbits of period $p \leq 10^3$ are already known (in the form of the logistic map $y_{n+1} = \mu y_n(1 - y_n)$) \cite{perrier2022scaling}. 
    Specifically, to obtain desired one-dimensional dynamics, we just have to choose the initial input value nearby target three data points and the corresponding feedback strength $\sigma_\text{fb}$, which is calculated by combining Eq.~\eqref{eq:kappa} with $\mu(\kappa) = \sqrt{-4\kappa + 1} + 1$, while making the trained map sufficiently close to a quadratic map around the target data.
    As mathematically verified in Sec.~\ref{subsubsec:small}, physical ELM with the analytic activation function $\phi$ and finite number of nodes $N < \infty$ can indeed accomplish such quadratic-like interpolation, provided that the corresponding Gram matrix $\hat{\Theta}$ (Eq.~\eqref{eq:Gram}) has full rank for sufficiently small input weight scaling $\sigma_w \ll 1$.
    Our thermodynamic limit analysis guarantees that $\hat{\Theta}$ would have full rank, at least if $\phi$ is a non-polynomial continuous function and $N$ is sufficiently large.
    However, we should be aware that if the target data points are in close proximity to each other, some trained networks may cause finite-size effects (e.g., when $\phi = \sin,\cos$, see Figs.~\ref{fig:finite_bf} and \ref{fig:finite_map}).
    In conclusion, LP3 can be considered an efficient and universal algorithm for programming all types of one-dimensional periodic orbits into an analog computer \cite{MacLennan2018}.
    This capability can be utilized to design the bifurcation of a physical system with desirable properties, such as durability in a high-radiation environment in which a computer-simulated attractor will break down \cite{akashi2022coupled}.
    We leave the further analysis of the control of embeddable attractors in physical systems for future work.

\begin{acknowledgments}
    We are grateful to Allen Hart for the fruitful discussions on attractor embedding in RC,
    to Ichiro Tsuda for highlighting the applicability of our theory to general datasets using Li--Yorke's condition,
    and to the RC seminar members for the stimulating discussions.
    K. N. is supported by JST CREST Grant Number JPMJCR2014 and by the Cross-ministerial Strategic Innovation Promotion Program (SIP) on “Integrated Health Care System” Grant Number JPJ012425.
\end{acknowledgments}

\appendix

\section{\label{appendix:variations}Variations in our network model}
    Here, we will present the corresponding kernel of the network with the output bias:
    \begin{equation*}
        f_N^{\text{bias}}(x)
        \equiv
        \frac{1}{\sqrt{N}}\sum_{i=1}^{N} W_i^{\text{out}} \phi\left(h_i(x)\right)
        + b^{\text{out}},
    \end{equation*}
    the network with input parameters drawn from uniform distribution:
    \begin{equation}
        \label{eq:uniform}
        W_i^{\text{in}}
        \sim
        {\cal U}(-1, 1)
        \,\,\text{and}\,\,
        b_i^{\text{in}}
        \sim
        {\cal U}(-1, 1),
    \end{equation}
    and the $L$-layer neural network with widths $n_\ell$ $(\ell=0,\ldots,L)$ \cite{jacot2018neural,lee2019wide}:
    \begin{equation}
        \label{eq:deep}
        \begin{gathered}
            \begin{cases}
                f_{\{n_\ell\}}^{\text{deep}}(x)
                &\equiv
                h^{(L+1)}(x),
                \\
                h^{(\ell+1)}(x)
                &
                \equiv
                \frac{\hat{\sigma}_w}{\sqrt{n_{\ell}}}W^{(\ell+1)} \alpha^{(\ell)}(x) + \hat{\sigma}_b b^{(\ell+1)},
                \\
                \alpha^{(0)}(x)
                &\equiv
                x,
                \\
                \alpha^{(\ell)}(x)
                &\equiv
                \phi(h^{(\ell)}(x)),
            \end{cases}
        \end{gathered}
    \end{equation}
    where $W^{(\ell+1)}\in\mathbb{R}^{n_{\ell+1} \times n_{\ell}}$ and $b^{(\ell+1)}\in\mathbb{R}^{n_{\ell+1}}$ are the weights and biases fundamentally initialized by an iid normal distribution;
    $\hat{\sigma}_w$ and $\hat{\sigma}_b$ are the constants governing the scales of weights and biases, respectively;
    $\phi$ is an element-wise activation function;
    and $n_0=n_{L+1}=1$.

    Adding the column vector $[1,\cdots,1] \in \mathbb{R}^{1\times|{\cal D}|}$ onto the matrix of hidden states ${\cal R}({\cal X})$,
    we obtained the kernel for the trained network with the output bias as follows:
    \begin{equation}
        \label{eq:NTK_bias}
        \begin{gathered}
            \Theta^{\text{bias}}(x,y)
            =
            \Theta(x,y)
            +
            1,
            \\
            \text{where}
            \quad
            \Theta(x,y)
            =
            \mathbb{E}
            \left[
                \phi(\omega x+\beta)\phi(\omega y+\beta)
            \right].
        \end{gathered}
    \end{equation}
    Eq.~\eqref{eq:NTK_bias} indicates that if $\Theta$ has full rank, then $\Theta^{\text{bias}}$ also has full rank; however, we note that the existence of $b^{\text{out}}$ does not ensure the full rank of $\Theta$.

    The kernel $\hat{\Theta}(x,y)$ (Eq.~\eqref{eq:TK}) with the input parameters drawn from the uniform distributions also converges in probability to $\Theta(x,y)$---that is, the expectation over random variables $\omega \sim {\cal U}(-\sigma_w, \sigma_w)\text{ and }\beta \sim {\cal U}(-\sigma_b, \sigma_b)$---within the limit $N\rightarrow\infty$ by the law of large numbers:
    \begin{equation}
        \label{eq:NTK_uniform}
        \begin{gathered}
            \Theta(x,y)
            =
            \frac{1}{4\sigma_w\sigma_b}
            \int_{-\sigma_w}^{\sigma_w}
            d\omega
            \int_{-\sigma_b}^{\sigma_b}
            d\beta \,
            \phi(\omega x + \beta)
            \phi(\omega y + \beta),
            \\
            \Theta^{\text{bias}}(x,y)
            =
            \Theta(x,y)
            +1.
        \end{gathered}
    \end{equation}
    As Eq.~\eqref{eq:NTK_uniform} is a definite integral, we can numerically compute NTK. Therefore, assuming that the matrix $\Theta$ has full rank, we can compute $f_\infty^*(x)$ for any input $x$.

    For deep neural networks (Eq.~\eqref{eq:deep}), we obtain two types of kernels ${\cal K}^{L+1}(x,y)$ (NNGP kernel) and $\Theta^{L+1}(x,y)$ (NTK) corresponding to the learning schemes: the readout-only training \cite{jacot2018neural,arora2019exact} and the lazy full training \cite{lee2019wide,arora2019exact}, respectively:
    \begin{gather*}
        {\cal K}^{\ell}(x,y)
        =
        \hat{\sigma}_w^2
        {\cal T}
        \left(
            \begin{bmatrix}
                {\cal K}^{\ell-1}(x,x) & {\cal K}^{\ell-1}(x,y) \\
                {\cal K}^{\ell-1}(x,y) & {\cal K}^{\ell-1}(y,y) \\
            \end{bmatrix}
        \right)
        +
        \hat{\sigma}_b^2,
        \\
        \begin{aligned}
            \Theta^{\ell}(x,y)
            &=
            \hat{\sigma}_w^2
            \Theta^{\ell-1}(x,y)
            \\
            &
            \times
            {\cal \dot{T}}
            \left(
                \begin{bmatrix}
                    {\cal K}^{\ell-1}(x,x) & {\cal K}^{\ell-1}(x,y) \\
                    {\cal K}^{\ell-1}(x,y) & {\cal K}^{\ell-1}(y,y) \\
                \end{bmatrix}
            \right)
            +
            {\cal K}^{\ell}(x,y),
        \end{aligned}
        \\
        \Theta^{1}(x,y)
        =
        {\cal K}^{1}(x,y)
        =
        \hat{\sigma}_w^2 xy + \hat{\sigma}_b^2,
    \end{gather*}
    where ${\cal T}$ and ${\cal \dot{T}}$ are functions from $2 \times 2$ positive semi-definite matrices:
    \begin{equation*}
        \Sigma
        \equiv
        \begin{bmatrix}
            \|x\|^2 & x \cdot y \\
            x \cdot y & \|y\|^2 \\
        \end{bmatrix}
    \end{equation*}
    to $\mathbb{R}$,
    defined for $\phi=\text{erf},\sin,\cos,\text{ReLU}$ as follows \cite{lee2019wide}:
    \begin{gather*}
        \begin{gathered}
            {\cal T}^{\text{erf}}(\Sigma)
            =
            \frac{2}{\pi}
            \arcsin
            \left(
                \frac{2 x \cdot y}{\sqrt{\left(1+2\|x\|^2\right)\left(1+2\|y\|^2\right)}}
            \right)
            ,
            \\
            {\cal \dot{T}}^{\text{erf}}(\Sigma)
            =
            \frac{4}{\pi}
            \det
            \left(
                I + 2\Sigma
            \right)^{-\frac{1}{2}}
            ,
        \end{gathered}
        \\
        \begin{gathered}
            \begin{aligned}
                &{\cal T}^{\sin}(\Sigma)
                \\
                &
                =
                \frac{1}{2}
                \left\{
                    e^{
                        -
                        \frac{1}{2}
                        \left(
                            \|x\|^2 - 2x \cdot y + \|y\|^2
                        \right)
                    }
                    -
                    e^{
                        -
                        \frac{1}{2}
                        \left(
                            \|x\|^2 + 2x \cdot y + \|y\|^2
                        \right)
                    }
                \right\}
                ,
            \end{aligned}
            \\
            \begin{aligned}
                &{\cal T}^{\cos}(\Sigma)
                \\
                &
                =
                \frac{1}{2}
                \left\{
                    e^{
                        -
                        \frac{1}{2}
                        \left(
                            \|x\|^2 - 2x \cdot y + \|y\|^2
                        \right)
                    }
                    +
                    e^{
                        -
                        \frac{1}{2}
                        \left(
                            \|x\|^2 + 2x \cdot y + \|y\|^2
                        \right)
                    }
                \right\}
                ,
            \end{aligned}
            \\
            {\cal \dot{T}}^{\sin}(\Sigma)
            =
            {\cal T}^{\cos}(\Sigma)
            ,
            \quad
            {\cal \dot{T}}^{\cos}(\Sigma)
            =
            {\cal T}^{\sin}(\Sigma)
            ,
        \end{gathered}
        \\
        \begin{gathered}
            \begin{gathered}
                {\cal T}^{\text{relu}}(\Sigma)
                =
                \frac{1}{2\pi}
                \|x\|
                \|y\|
                \left\{
                    \sqrt{1-\cos^2\psi}
                    +
                    (\pi - \psi)
                    \cos\psi
                \right\}
                ,
                \\
                {\cal \dot{T}}^{\text{relu}}(\Sigma)
                =
                \frac{1}{2\pi}
                (\pi - \psi)
                ,
            \end{gathered}
            \\
            \text{where}
            \quad
            \psi
            \equiv
            \arccos
            \frac{x \cdot y}{\|x\|\|y\|}.
        \end{gathered}
    \end{gather*}
    Optimizing weights and biases by minimizing mean squared error loss ${\cal L}=\frac{1}{2}\|f({\cal X})-{\cal Y}\|^2$, the (continuous-time) dynamics of the (linearized) network output $f_\infty^{\text{lin}}$ within its thermodynamic limit $n_1,\ldots,n_L\rightarrow\infty$ is described as follows:
    \begin{equation*}
        \label{eq:finf_deep}
        \begin{aligned}
            &f_\infty^{\text{lin}}(x,t)
            =
            K(x,{\cal X})
            K^{-1}
            \left(
                I - e^{- \eta K t}
            \right){\cal Y}
            \\
            &+
            f_\infty^{\text{deep}}(x)
            -
            K(x,{\cal X})
            K^{-1}
            \left(
                I - e^{- \eta K t}
            \right)
            f_\infty^{\text{deep}}({\cal X})
        \end{aligned}
    \end{equation*}
    where $K={\cal K}^{L+1}$ (readout-only training) or $K=\Theta^{L+1}$ (lazy full training), $t$ is the time variable for the training dynamics, and $\eta$ is the learning rate.
    With a small initial output $f_\infty^{\text{deep}}(x) \approx 0$, the network output asymptotically becomes a kernel regression predictor \cite{jacot2018neural,lee2019wide,arora2019exact}:
    \begin{equation}
        \label{eq:finf_deep_kr}
        f_\infty^{*,\text{deep}}(x)
        \approx
        \lim_{t\rightarrow\infty}
        f_\infty^{\text{lin}}(x,t)
        \approx
        K(x,{\cal X})
        K^{-1}{\cal Y}.
    \end{equation}
    In readout-only training, where $K={\cal K}^{L+1}$, the approximation becomes exact when the initial readout weights are set to zero, which is equivalent to performing least square regression \cite{ali2019continuous,advani2020high}.
    Note that if $L = 1$ (one-layer) and $\hat{\sigma}_w = \hat{\sigma}_b = \sigma$, then the NNGP kernel ${\cal K}^{2}(x,y)$ coincides with the kernel $\Theta(x,y)$ in Eq.~\eqref{eq:NTK_bias},
    except the multiplier $\sigma^2$, which will diminish by the multiplication of $K(x,{\cal X})$ and $K^{-1}$ in Eq.~\eqref{eq:finf_deep_kr}.

\section{\label{appendix:formulas}Partial derivatives of NTKs}
    The formulas for $\frac{\partial \Theta}{\partial x}(x,y)$ when $\phi=\text{erf},\sin,\cos,\text{ReLU}$ and $\sigma_w=\sigma_b=\sigma$ are as follows:
    \begin{equation}
        \begin{aligned}
            \frac{\partial\Theta^{\text{erf}}}{\partial x}
            &(x,y)
            =
            \frac{4\sigma^2}{\pi\left[1+2\sigma^2(1+x^2)\right]}
            \\
            \times
            &\frac{y - 2\sigma^2(x-y)}{\sqrt{1+2\sigma^2(2 + x^2 + y^2) + 4\sigma^4(x-y)^2}}
            ,
        \end{aligned}
    \end{equation}
    \begin{gather}
        \begin{aligned}
            \frac{\partial\Theta^{\sin}}{\partial x}(x,y)
            =
            -\frac{\sigma^2}{2}
            &
            \left\{
                (x-y)e^{-\frac{\sigma^2}{2}(x-y)^2}
            \right.
            \\
            &
            \left.
                -
                (x+y)e^{-\frac{\sigma^2}{2}(x+y)^2-2\sigma^2}
            \right\}
            ,
        \end{aligned}
        \\
        \begin{aligned}
            \frac{\partial\Theta^{\cos}}{\partial x}(x,y)
            =
            -\frac{\sigma^2}{2}
            &
            \left\{
                (x-y)e^{-\frac{\sigma^2}{2}(x-y)^2}
            \right.
            \\
            &
            \left.
                +
                (x+y)e^{-\frac{\sigma^2}{2}(x+y)^2-2\sigma^2}
            \right\}
            ,
        \end{aligned}
        \\
        \begin{gathered}
            \frac{\partial\Theta^{\text{relu}}}{\partial x}(x,y)
            =
            \frac{\sigma^2}{2\pi}
            \left\{
                \frac{x|x-y|}{1+x^2}
                +
                (\pi - \psi)y
            \right\}
            ,
            \\
            \text{where}
            \quad
            \psi
            \equiv
            \arccos
            \frac{1+xy}{\sqrt{\left(1+x^2\right)\left(1+y^2\right)}}.
        \end{gathered}
    \end{gather}
    To calculate the relative error $e$ (Eq.~\eqref{eq:error}) for $\phi=\text{erf},\sin,\cos,\text{ReLU}$, with $\sigma_b = 1$ in the thermodynamic limit (Fig.~\ref{fig:confirm}), we utilized the following formulas:
    \begin{gather}
        \begin{aligned}
            &
            \left.
            \frac{d^nf_\infty^*}{dx^n}(x)
            \right|_{\sigma_b=1}
            \\
            &=
            \sigma_w^n
            \frac{\partial^n k}{\partial x^n}
            (\sigma_w x,\sigma_w {\cal X}) 
            \left[
                k(\sigma_w {\cal X},\sigma_w {\cal X})
            \right]^{-1}
            {\cal Y},
        \end{aligned}
        \\
        \frac{\partial^n k}{\partial x^n}
        (x,y)
        \equiv
        \left.
            \frac{\partial^n \Theta}{\partial x^n}
            (x,y)
        \right|_{\sigma=1}
        ,
        \\
        \frac{\partial^2\Theta^{\text{relu}}}{\partial x^2}(x,y)
        =
        \frac{\sigma^2}{2\pi}
        \cdot
        \frac{2|x-y|}{(1+x^2)^2}
        .
    \end{gather}

\section{\label{appendix:numerical}Numerical analysis of learned periods and a comparative study of learning period $n=1,2,3,\ldots$}

    \begin{figure*}
        \includegraphics[width=\textwidth]{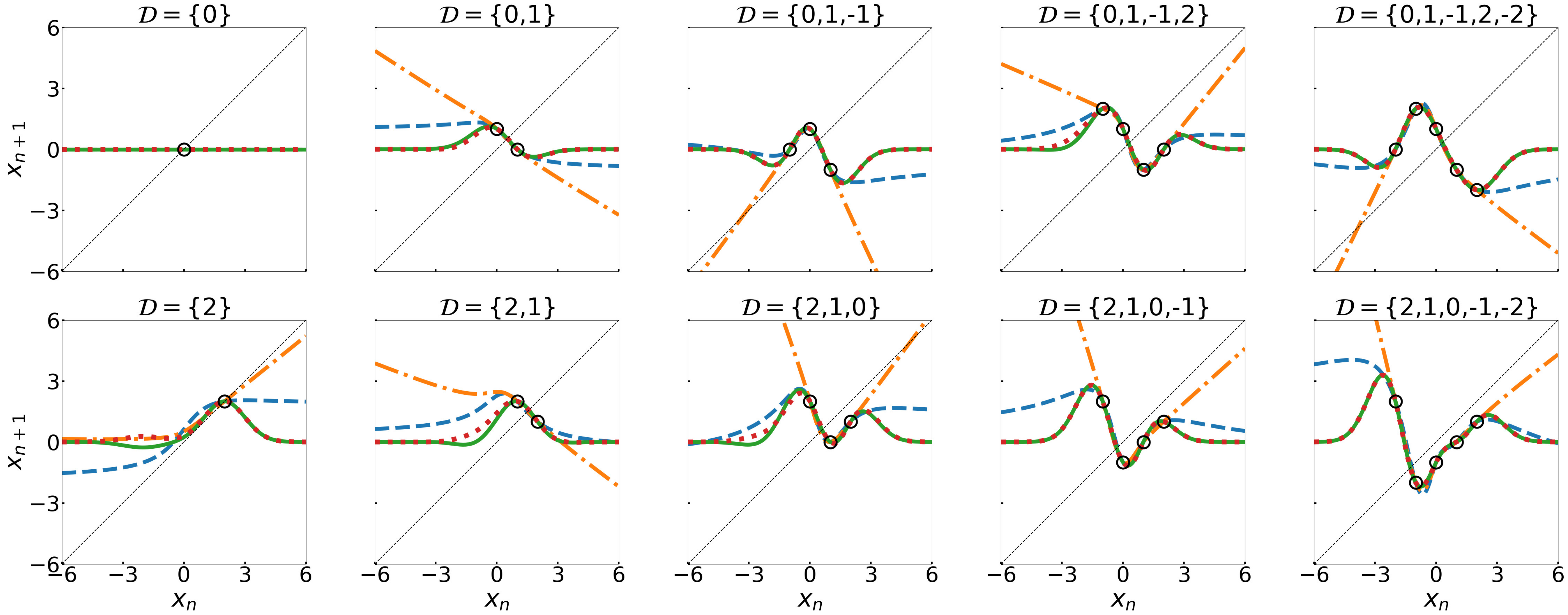}
        \caption{
            Trained maps $f_\infty^*$ in learning period $n=1,2,3,4,5$, with $\sigma=1.0$ for $\phi=\text{erf}$ (blue line), $\sin$ (green line), $\cos$ (red line), and $\text{ReLU}$ (orange line).
            Even in learning period $n=1,2$ (the two leftmost columns), where there is only one type of ${\cal D}$, $f_\infty^*$ depends on the value of the target data.
            Increasing $n$ explodes the number of types of ${\cal D}$, resulting in the strong dependence of $f_\infty^*$ on the ordering of periodic points in ${\cal D}$, as can be seen in the case of learning period $n=5$ (the rightmost column).
        }
        \label{fig:period_n_trained_map}
    \end{figure*}

    \begin{figure*}
        \includegraphics[width=\textwidth]{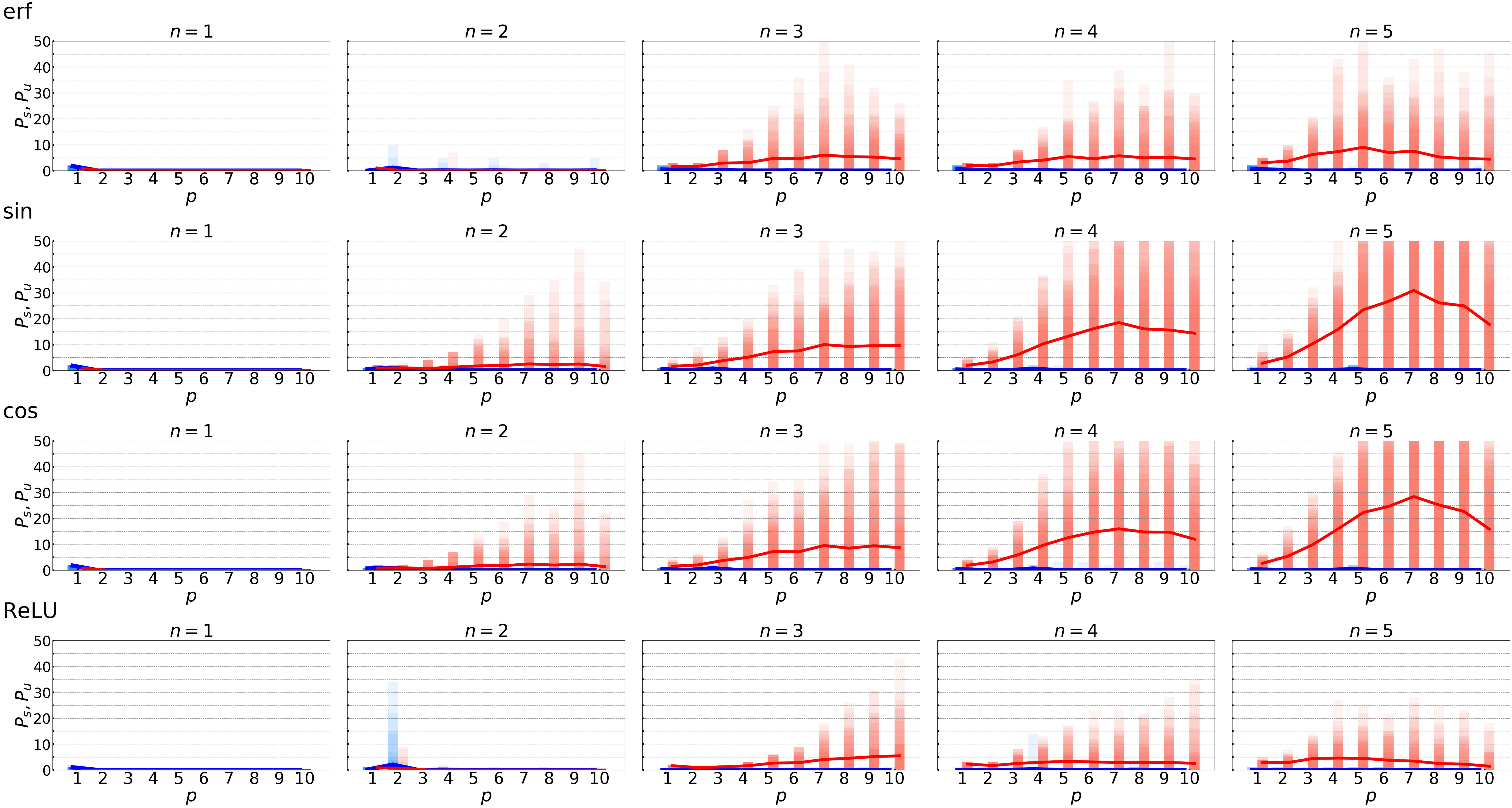}
        \caption{
            Distribution of the learned periods in learning period $n$, with $\sigma=1.0$.
            Histograms for 100 different realizations of target data ${\cal D}$ are overlaid for each $n$.
            The blue and red bins indicate the number of periodic orbits of period $p$ with $\lambda_p < 0$ (stable, $P_s$) and $\lambda_p > 0$ (unstable, $P_u$), respectively.
            The solid lines show the average distributions of the learned periods in learning period $n$.
            Periodic orbits of $\lambda_p = 0$ were not detected in this setting.
        }
        \label{fig:period_n_distribution}
    \end{figure*}

    \begin{figure*}
        \includegraphics[width=\textwidth]{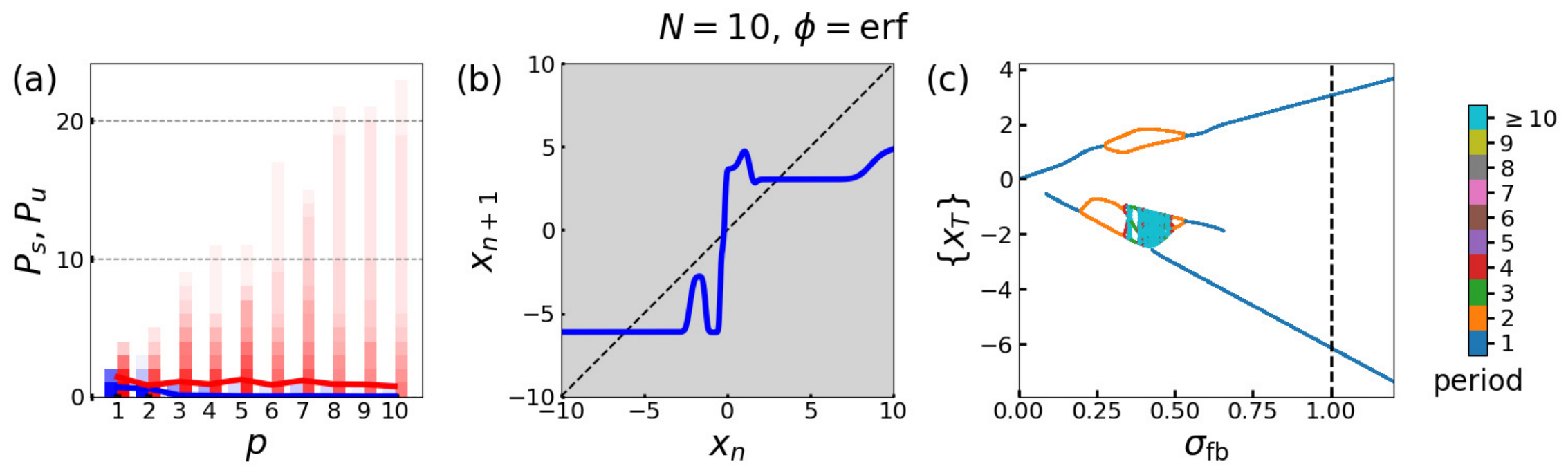}
        \caption{
            Latently acquired periods in the random neural network $f_N$ with the random network weights $W^{\text{in}}$ and $W^{\text{out}}$ generated from ${\cal N}(0,5.0)$.
            (a) Distribution of periods.
            (b) Example of $f_N$.
            (c) Externalization $\sigma_\text{fb}f_N$, calculated with $-10 \leq x_0 \leq 10$ and $T=10^3$.
            Histograms for 100 different realizations of the network weights $W^{\text{in}}$ and $W^{\text{out}}$ are overlaid.
            The blue and red bins indicate the number of stable ($P_s$) and unstable ($P_u$) periodic orbits of period $p$, respectively.
            The solid lines in the left panel show the average distributions of the periods.
        }
        \label{fig:randomNN}
    \end{figure*}

    To investigate the learned periods, including unstable ones, we solve the following nonlinear equation:
    \begin{equation}
        \label{eq:numerical}
        (f_\infty^*)^{p}(x_p) - x_p = 0
        ,
    \end{equation}
    using the MATLAB $\mathtt{fsolve}$ command.
    We uniformly chose $10^3$ initial points from the intervals
    $[-10.0, 10.0]$ for Figs.~\ref{fig:1dbifurcation}(a) and \ref{fig:externalization}(b), and $[-100.0,100.0]$ for Figs.~\ref{fig:period_n_distribution} and ~\ref{fig:randomNN}, to numerically solve this equation.
    To count the number of periodic orbits of period $p$, we used the absolute tolerance $10^{-2}$ to exclude the points belonging to the same periodic point, periodic orbit, and the periodic point of period $p' < p$ from the numerical solutions of Eq.~\eqref{eq:numerical}.

    We then considered learning period $n=1,2,3,4,5$ (see Fig.~\ref{fig:period_n_trained_map} for the examples of the trained maps) to investigate how $n$ affects the learned periods. 
    Fig.~\ref{fig:period_n_distribution} shows the average numbers of learned periods $p=1,2,\ldots,10$ in the map $f_\infty^*$ through learning period $n$ with a randomly drawn ${\cal D}$ from the interval $[-10.0,10.0]$.
    We also calculated the stability of the detected periodic orbit by computing $\lambda_p \equiv \ln|\frac{d}{dx}(f_\infty^*)^{p}(x_p)|$; we considered the periodic orbits of $\lambda_p < 0$ stable, and those of $\lambda_p > 0$, unstable.
    It is important to note that the average distribution is affected by the tolerance $10^{-2}$ and the choice of the initial points in numerical calculations.
    We observed that, regardless of the choice of $\phi$, the number of unstable periods tends to increase dramatically after $n=3$.
    This phenomenon may correspond to the fact that there always exists an appropriate ordering of period $n\geq 3$ that induces all periods (a Štefan sequence \cite{burns2011sharkovsky} of length 3), as discussed in Sec.~8 in Ref.~\cite{burns2011sharkovsky}.
    Meanwhile, learning period two leads to non-trivial phenomena, depending on the choice of ${\cal D}=\{a,b\}$ and $\phi$.
    For $\phi=\text{ReLU}$, $f_\infty^*$ becomes similar to $f(x)=-x+\alpha\,(\alpha\in\mathbb{R})$ for some choices of ${\cal D}$, resulting in a large amount of period-2 orbits (see also Fig.~\ref{fig:period_n_trained_map}).
    For $\phi=\sin,\cos$, some choices of ${\cal D}$ provide a period-three orbit in $f_\infty^*$, thereby inducing Li--Yorke chaos.

    Similarly, we could investigate the intrinsic periodic orbits in random neural networks without modifying their connectivity through learning.
    We found that certain random neural networks inherently possess period three (see Fig.~\ref{fig:randomNN}).
    This observation is consistent with the findings in Ref.~\cite{ishihara2005magic}, which suggest that chaos can arise in layered random neural networks with a small number of neurons.

\section{\label{appendix:proof}Proof of Theorem~\ref{prop:sym_tc}}

    Applying $b=-a$, $\Theta(x,y)=\Theta(y,x)$, and $\Theta(x,y)=\Theta(-x,-y)$ to Eq.~\eqref{eq:finf_p3}, $\left.f_\infty^*(x) \right|_{{\cal D}=\{a,-a,c\}}$ is given by
    \begin{widetext}
        \begin{equation}
            \label{eq:finf_p3_bis-a}
            \begin{aligned}
                \left.f_\infty^*(x) \right|_{{\cal D}=\{a,-a,c\}}
                =
                -\frac{a}{|\Theta|}
                &
                \left[
                    \Theta(x,a)
                    \left\{
                        \Theta(a,a)\Theta(c,c)-\Theta(c,-a)^2
                    \right\}
                \right.
                \\
                    &+
                    \Theta(x,-a)
                    \left\{
                        \Theta(c,-a)\Theta(c,a)-\Theta(a,-a)\Theta(c,c)
                    \right\}
                \\
                    &+
                \left.
                    \Theta(x,c)
                    \left\{
                        \Theta(a,-a)\Theta(c,-a)-\Theta(a,a)\Theta(c,a)
                    \right\}
                \right]
                \\
                +
                \frac{c}{|\Theta|}
                &
                \left[
                    \Theta(x,a)
                    \left\{
                        \Theta(c,-a)\Theta(c,a)-\Theta(a,-a)\Theta(c,c)
                    \right\}
                \right.
                \\
                    &+
                    \Theta(x,-a)
                    \left\{
                        \Theta(c,c)\Theta(a,a)-\Theta(c,a)^2
                    \right\}
                \\
                    &+
                \left.
                    \Theta(x,c)
                    \left\{
                        \Theta(a,-a)\Theta(c,a)-\Theta(a,a)\Theta(c,-a)
                    \right\}
                \right]
                \\
                +
                \frac{a}{|\Theta|}
                &
                \left[
                    \Theta(x,a)
                    \left\{
                        \Theta(a,-a)\Theta(c,-a)-\Theta(a,a)\Theta(c,a)
                    \right\}
                \right.
                \\
                    &+
                    \Theta(x,-a)
                    \left\{
                        \Theta(a,-a)\Theta(c,a)-\Theta(a,a)\Theta(c,-a)
                    \right\}
                \\
                    &+
                \left.
                    \Theta(x,c)
                    \left\{
                        \Theta(a,a)^2-\Theta(a,-a)^2
                    \right\}
                \right],
            \end{aligned}
        \end{equation}
        \begin{equation}
            \label{eq:det_p3_bis-a}
            \begin{aligned}
                |\Theta|
                =
                \Theta(a,a)^2\Theta(c,c)
                +
                2\Theta(a,-a)\Theta(c,a)\Theta(c,-a)
                -
                \Theta(a,a)
                \left\{
                    \Theta(c,a)^2
                    +
                    \Theta(c,-a)^2
                \right\}
                -
                \Theta(c,c)\Theta(a,-a)^2.
            \end{aligned}
        \end{equation}
    \end{widetext}

    We note that $|\Theta|$ (Eq.~\eqref{eq:det_p3_bis-a}) is invariant under the sign change of $a$ or $c$ ($a \rightarrow -a$ or $c \rightarrow -c$).
    Now, let us consider the another type of LP3 (${\cal D}={\{-a,a,c\}}$)---that is, the sign change of $a$:
    \begin{widetext}
        \begin{equation}
            \label{eq:finf_p3_othertype}
            \begin{aligned}
                \left.f_\infty^*(x) \right|_{{\cal D}=\{-a,a,c\}}
                =
                \frac{a}{|\Theta|}
                &
                \left[
                    \Theta(x,-a)
                    \left\{
                        \Theta(a,a)\Theta(c,c)-\Theta(c,a)^2
                    \right\}
                \right.
                \\
                    &+
                    \Theta(x,a)
                    \left\{
                        \Theta(c,a)\Theta(c,-a)-\Theta(a,-a)\Theta(c,c)
                    \right\}
                \\
                    &+
                \left.
                    \Theta(x,c)
                    \left\{
                        \Theta(a,-a)\Theta(c,a)-\Theta(a,a)\Theta(c,-a)
                    \right\}
                \right]
                \\
                +
                \frac{c}{|\Theta|}
                &
                \left[
                    \Theta(x,-a)
                    \left\{
                        \Theta(c,a)\Theta(c,-a)-\Theta(a,-a)\Theta(c,c)
                    \right\}
                \right.
                \\
                    &+
                    \Theta(x,a)
                    \left\{
                        \Theta(c,c)\Theta(a,a)-\Theta(c,-a)^2
                    \right\}
                \\
                    &+
                \left.
                    \Theta(x,c)
                    \left\{
                        \Theta(a,-a)\Theta(c,-a)-\Theta(a,a)\Theta(c,a)
                    \right\}
                \right]
                \\
                -
                \frac{a}{|\Theta|}
                &
                \left[
                    \Theta(x,-a)
                    \left\{
                        \Theta(a,-a)\Theta(c,a)-\Theta(a,a)\Theta(c,-a)
                    \right\}
                \right.
                \\
                    &+
                    \Theta(x,a)
                    \left\{
                        \Theta(a,-a)\Theta(c,-a)-\Theta(a,a)\Theta(c,a)
                    \right\}
                \\
                    &+
                \left.
                    \Theta(x,c)
                    \left\{
                        \Theta(a,a)^2-\Theta(a,-a)^2
                    \right\}
                \right].
            \end{aligned}
        \end{equation}
    \end{widetext}

    Applying the transformation $x \rightarrow -x$ and $c \rightarrow -c$ to Eq.~\eqref{eq:finf_p3_othertype}, we obtain
    \begin{equation}
        \tag{\ref{eq:sym_tc}}
        \left. f_\infty^*(-x) \right|_{{\cal D}=\{-a,a,-c\}}
        =
        -
        \left. f_\infty^*(x) \right|_{{\cal D}=\{a,-a,c\}},
    \end{equation}
    which is what we wanted to prove.

% \nocite{*}

\bibliography{LP3}

\end{document}